\documentclass{article}
\PassOptionsToPackage{numbers,sort&compress}{natbib}
\usepackage[preprint]{neurips_2026}
\usepackage[utf8]{inputenc} % allow utf-8 input
\usepackage[T1]{fontenc}    % use 8-bit T1 fonts
\usepackage{hyperref}       % hyperlinks
\usepackage{url}            % simple URL typesetting
\usepackage{booktabs}       % professional-quality tables
\usepackage{amsfonts}       % blackboard math symbols
\usepackage{nicefrac}       % compact symbols for 1/2, etc.
\usepackage{microtype}      % microtypography
\usepackage{xcolor}         % colors
\usepackage{graphicx}
\usepackage{amsmath}
\usepackage{multirow}
\usepackage{array}
\usepackage{subcaption}
\usepackage{comment}

\usepackage{tikz}
\usepackage{amsmath}
\usepackage{amsfonts}
\usetikzlibrary{shapes.geometric, arrows.meta, positioning, fit, backgrounds, calc}

\definecolor{FrozenBlue}{HTML}{2C3E50}
\definecolor{TrainYellow}{HTML}{F39C12}
\definecolor{AuxGray}{HTML}{BDC3C7}
\definecolor{GeluOrange}{HTML}{E67E22}
\title{Resource-Aware Parameter-Efficient Model Adaptation for Onboard High-Dimensional Data}

\author{%
  \textbf{Qiyang Zhang}\textsuperscript{1}\hspace{1em}
  \textbf{Xinhao Li}\textsuperscript{2}\hspace{1em}
  \textbf{Lei Shi}\textsuperscript{3}\hspace{1em}
  \textbf{Zheng Lin}\textsuperscript{4}\hspace{1em}
  \textbf{Jinfeng Wen}\textsuperscript{1}\hspace{1em}\\[0.15em]
  \textbf{Ao Zhou}\textsuperscript{1}\hspace{1em}  
  \textbf{Shangguang Wang}\textsuperscript{1}
  \\[0.35em]
  \normalfont\small
  \textsuperscript{1}Beijing University of Posts and Telecommunications
  \qquad
  \textsuperscript{2}Wuhan University
  \\[0.10em]
  \textsuperscript{3}Communication University of China
  \qquad
  \textsuperscript{4}University of Luxembourg
  \\[0.30em]
  \normalfont\small
  \texttt{\{qiyzhang,jinfeng.wen,aozhou,sgwang\}@bupt.edu.cn}
  \\[0.10em]
  \texttt{xhl17866703382@163.com,\quad
  leiky\_shi@cuc.edu.cn,\quad
  zhenglin@ieee.org}
}

\begin{document}

\maketitle

\begin{abstract}
Onboard satellite models often require frequent updates, but the weights adapted to earlier data distributions can quickly become outdated. However, updating large-scale model parameters in orbit presents significant challenges due to the limited uplink bandwidth of Low Earth Orbit (LEO) satellite systems, particularly for hyperspectral satellite imagery, where high-dimensional spectral–spatial inputs lead to increased model size and update costs. 
Existing full fine-tuning methods are thus expensive to retrain and difficult to deploy under strict communication constraints.
To address this challenge, we propose NE-LoRA, a parameter-efficient adaptation framework for bandwidth-constrained onboard hyperspectral model updates. NE-LoRA combines a primary low-rank branch with a nonlinear auxiliary branch to capture both global update trends and complex spectral–spatial variations. 
Additionally, we introduce a differentiated training strategy for multi-matrix adapters, motivated by the asymmetric initialization and gradient dynamics of different adapter matrices.
Experiments on four hyperspectral datasets and three representative backbone models demonstrate that NE-LoRA consistently outperforms LoRA-based baselines and remains competitive with, and in several cases superior to, full fine-tuning. Across the evaluated settings, NE-LoRA updates only a small fraction of the total parameters on average while preserving low deployment overhead, offering a favorable accuracy–communication trade-off for onboard hyperspectral adaptation.
\end{abstract}

\section{Introduction}
Low Earth Orbit (LEO) satellites play a crucial role as key edge nodes in emerging LEO satellite networks (LSNs), offering vital sensing and communication capabilities for modern networked applications. Use cases such as machine learning (ML)-driven Earth monitoring and emergency response \cite{chen2020cascading, lin2026leo,lang2023high, zhang2026satsfl,li2024satguard, toker2022dynamicearthnet,lin2025fedsn} are expected to become increasingly important in the future satellite Internet.

However, a significant challenge to practical onboard intelligence is the absence of efficient model updating mechanisms in conventional satellite architectures.
Without timely updates, onboard models cannot struggle to adapt to evolving data distributions or shifting application demands \cite{norris2016evidence, xi2022beyond, yang2013role}, leading to a gradual degradation or even failure of application functionalities \cite{hort2021survey}.
When updates are required, the common solution is to retrain and redistribute model weights from scratch, which is time-consuming and environmentally unsustainable.
For example, \cite{ekelund2024ai} reports that uploading updated neural network parameters can take anywhere from several minutes to several hours, severely limiting the practicality of frequent onboard fine-tuning.
This challenge is expected to worsen as model size and update frequency continue to increase \cite{xu2024fwdllm}.
In hyperspectral satellite systems, the difficulty is further exacerbated by the high resolution and complex spectral–spatial information, which result in high-dimensional inputs, larger models, and increased update costs.
Consequently, efficient, communication-aware model updating is crucial for maintaining accuracy and uplink efficiency.

Updating hyperspectral satellite models introduces  three major challenges: (i) \textit{Diverse models and limited onboard resources.} 
Onboard satellite models require periodic updates to sustain performance and generalization. However, limited computation, memory, and energy make frequent or model-specific adaptation difficult, with direct onboard training often being infeasible.
(ii) \textit{Constrained bandwidth and intermittent connectivity.} Updated parameters must be transmitted over narrow bandwidth within short satellite--ground contact windows, rendering full-model transmission impractical.
(iii) \textit{High-dimensional and information-rich hyperspectral feature.} The large number of spectral bands and complex spectral--spatial structure of hyperspectral imagery significantly increase both model size and adaptation cost.
Existing methods do not fully address these three challenges simultaneously.
A clear gap remains in developing model update approaches that satisfy application requirements, communication constraints, and hyperspectral model characteristics.
This gap motivates the design of a parameter-efficient and communication-aware adaptation framework for hyperspectral satellite model updating.

\begin{figure*}[t]
    \centering
    
    % 左边：完整图
    \begin{minipage}[c]{0.42\textwidth}
        \centering
        \includegraphics[width=0.98\linewidth]{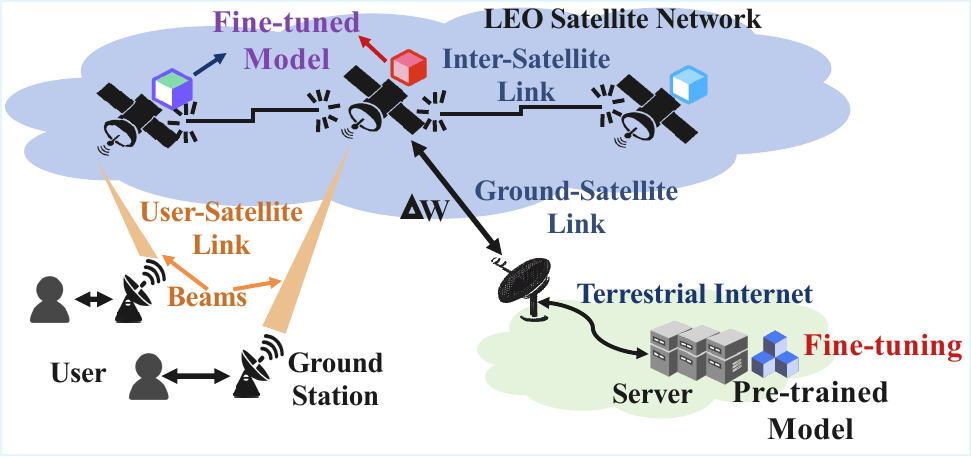}
        \captionof{figure}{A high-level architecture of today’s LSNs.}
        \label{fig:main_architecture}
    \end{minipage}
    \hfill
    % 右边：一个整体图，内部含两个子图
    \begin{minipage}[c]{0.54\textwidth}
        \centering
        
        \begin{subfigure}[c]{0.48\linewidth}
            \centering
            \includegraphics[width=\linewidth,height=3.0cm]{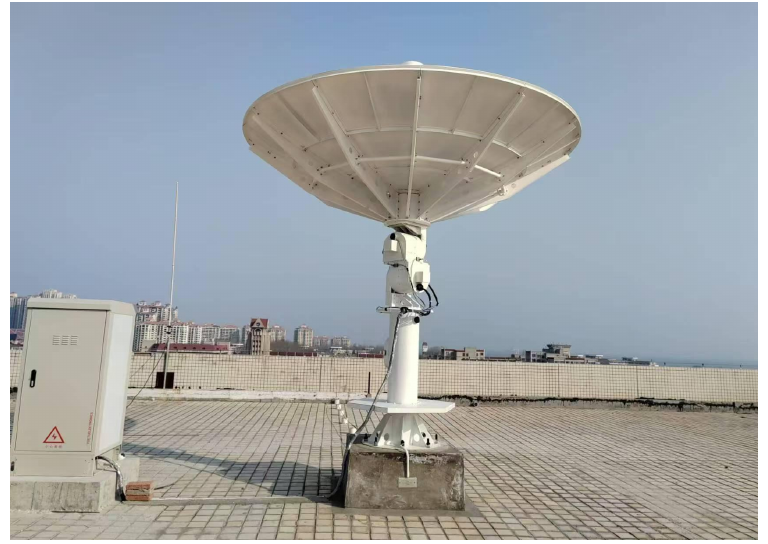}
            \caption{Experimental setup of data collection.}
            \label{fig:exp_setup}
        \end{subfigure}
        \hfill
        \begin{subfigure}[c]{0.48\linewidth}
            \centering
            \includegraphics[width=\linewidth,height=3.0cm]{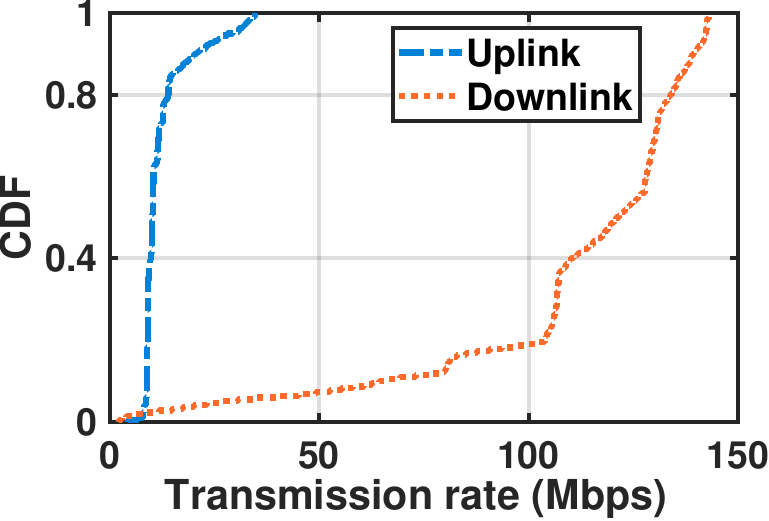}
            \caption{CDF of uplink/downlink transmission rates.}
            \label{fig:cdf_rate}
        \end{subfigure}
        
        \captionof{figure}{Empirical measurements on an operational satellite.}
        \label{fig:measurement_data}
    \end{minipage}
    
    \vspace{-0.5em}
\end{figure*}

Research into onboard model updates is still in its early stage.
%Existing methods developed for terrestrial environments can be broadly grouped into three categories: (i) Incremental learning approaches (e.g., few-shot distillation) \cite{du2023incremental, li2017learning}, which mitigate catastrophic forgetting during model updates.
%However, these methods typically overlook communication efficiency and often require recurrent access to historical data, making them ill-suited to bandwidth-constrained satellite settings. (ii) Lightweight compression methods (e.g., knowledge distillation \cite{meng2023distillation}, pruning \cite{cho2023pdp}) which reduce model size or training cost.
%While effective in some settings, these methods often depend on sufficiently annotated data or may weaken spectral dependencies that are important for hyperspectral classification, thereby limiting their applicability to satellite scenarios. (iii) Low-Rank Adaptation (LoRA) \cite{hayou2024lora+, hu2022lora}, which adapts pretrained models using a small number of trainable parameters and is therefore more promising for communication-constrained model updating.
Most existing LoRA variants rely on linear low-rank update structures, which may be insufficient for capturing the complex nonlinear spectral--spatial interactions required for hyperspectral analysis \cite{Yang2024LowRankAF, 10903446, qiang2024bilorabileveloptimizationframework, jung2025graloragranularlowrankadaptation, dong2025aurora}.
As a result, hyperspectral satellite model updating requires a framework that balances onboard accuracy, parameter efficiency, and communication cost.

To address these limitations, we propose NE-LoRA, a nonlinear-enhanced and differentiated adaptation framework for hyperspectral satellite model updating.
NE-LoRA augments standard low-rank adaptation with a Gaussian Error Linear Unit (GELU)-activated auxiliary branch to improve nonlinear spectral--spatial modeling while preserving the low-rank structure.
Unlike existing nonlinear LoRA variants developed for general vision tasks, our framework is specifically designed for bandwidth-constrained onboard hyperspectral adaptation, where upload cost a critical factor, and spectral--spatial correlations differ substantially from conventional RGB imagery.
In addition, while standard LoRA applies a uniform learning rate to all adapter matrices, the asymmetric initialization and gradient dynamics of different matrices can lead to inefficient or unstable optimization.
This issue is particularly problematic in hyperspectral models, where large-proportion modules account for a significant portion of the overall parameter update and often exhibit highly asymmetric gradient behaviors.
Without explicit control, uniform learning rates may either over-amplify noisy updates or suppress important adaptation signals.
To overcome this issue, we introduce a differentiated training strategy for multi-matrix structures.
Specifically, we extend the standard LoRA update by incorporating: (i) a scaling factor to amplify the output-side modules with weak gradient signals, and (ii) a reduction factor to stabilize large proportion modules by suppressing excessive updates.

As illustrated in Fig. \ref{fig:main_architecture}, the update process begins with the satellite downlinking a small set of newly acquired samples to the ground station.
The ground station then trains the dual-branch adapters using its relatively abundant computational resources.
Before uplink transmission, the learned update is folded into an equivalent low-rank form so that the deployed upload cost remains in the same order as standard LoRA.
Upon reception, the satellite integrates the transmitted update into the original model without additional adaptation overhead, thus balancing ground-side flexibility and onboard efficiency.

Our main contributions are summarized as follows:
\begin{itemize}
\item We study parameter-efficient model adaptation for bandwidth-constrained onboard hyperspectral applications, a setting where spectral–spatial complexity and limited uplink capacity jointly make onboard model updates difficult.
\item We propose NE-LoRA, a dual-branch adaptation framework that augments standard low-rank updates with a nonlinear auxiliary branch, enabling more expressive modeling of hyperspectral spectral–spatial variations while retaining lightweight adaptation. 
%\item We introduce a differentiated training strategy for multi-matrix adapters, which accounts for asymmetric initialization and gradient dynamics across adapter matrices and improves optimization stability.

\item We introduce a differentiated training strategy for multi-matrix adapters, which mitigates the issue that large-proportion modules account for a significant portion of the total parameter update, and a uniform learning rate either over-amplifies noisy updates or suppresses important adaptation signals, thereby improving optimization stability.

%\textcolor{blue}{Change1(lxh): Asymmetric initialization among adapter matrices is not our contribution; our main innovation lies in the differentiated strategy tailored for large-proportion modules.}
\item Extensive evaluation on four hyperspectral datasets and three backbone architectures show that NE-LoRA consistently improves over LoRA-based baselines and provides a favorable accuracy–communication trade-off under deployment constraints.
%\item We have created a public repository\footnote{The data and code will be made publicly available upon paper acceptance.} with the data and code from this work to facilitate reproducibility and support future research.
\end{itemize}

%The remainder of the paper is organized as follows. Section 2 introduces the background and related work. Section 3 the proposed methodology. Section 4 describes the experimental setup and evaluation results. Section~5 concludes the paper.

\section{Background \& Related Work}
\subsection{Background}

To characterize the communication constraints of onboard model updating, we consider a real-world LEO satellite computing platform and measure its downlink and uplink rates. 
Fig.~\ref{fig:measurement_data}(a) illustrates the experimental setup used for data collection.
We employ a widely used network measurement tool (i.e., Iperf \cite{29}) to record downlink and uplink throughput, and report the corresponding cumulative distribution functions (CDFs) in Fig.~\ref{fig:measurement_data}(b).
The measurements show a clear communication asymmetry: the average downlink rate is close to 100 Mbps, whereas the average uplink rate is only about 12 Mbps.
For onboard model updating, this limited uplink capacity becomes a critical bottleneck for parameter delivery from the ground station to satellites. 
For example, transmitting a VGG-16 model (528 MB) from the ground requires approximately 5.9 minutes, which is too long to distribute updated models to all satellites within a single satellite--ground contact window.
These results highlight the stringent communication constraints faced by model updates.

Beyond uplink bottlenecks, onboard models are also constrained by limited computational resources, which restrict the complexity of feasible update algorithms. Frequent in-orbit updates, especially for large or high-dimensional models, can quickly exceed available compute and memory budget.
Taken together, limited uplink bandwidth and constrained onboard computation make efficient model updating a critical challenge for satellite intelligence.

Satellite constellations, consisting of numerous broadband satellites equipped with user-satellite links (USLs) \cite{2988}, extend the boundary of today’s Internet and support emerging networked services at a global scale.
Many of these satellites are equipped with hyperspectral sensors, which capture reflected signals across a large number of contiguous spectral bands and therefore produce high-dimensional spectral--spatial data. Unlike conventional RGB imagery with only three channels, hyperspectral imagery typically contains tens to hundreds of bands, substantially increasing the number of model input channels and the associated adaptation cost. It is note that the number of spectral bands highlights the high spectral dimensionality is a defining characteristic of hyperspectral data. For instance,  on the WHU-LK dataset, there is 270 bands.
This characteristic further amplifies the difficulty of efficient onboard model updating.

\subsection{Related Work}

%This subsection provides recently published and related works on parameter-efficient fine-tuning techniques. In general, these techniques aim to adapt pre-trained models to new tasks while reducing training costs. In the literature, these methods are broadly divided into three categories:
Fine-tuning aims to adapt pretrained models to new tasks while controlling computational and training costs. Existing methods related to model updating can be broadly grouped into three categories.

The first category includes approaches for mitigating catastrophic forgetting, such as learning without forgetting \cite{li2017learning} and few-shot distillation \cite{lin2024revisiting}.
These methods aim to preserve previously acquired knowledge during model updates.
For hyperspectral image classification, however, applying such methods often requires retaining feature representations, outputs, or representative samples from previous tasks.
Due to the high dimensionality of hyperspectral inputs, storing and repeatedly accessing such information can quickly exceed the storage and communication budget available in satellite systems.
As a result, although these methods are effective for continual adaptation, they are not designed for onboard updating and often rely on recurrent access to historical data.

The second category includes lightweight compression approaches, such as distillation and pruning, which aim to reduce model size or training cost.
For example, Meng et al.~\cite{meng2023distillation} proposed a distillation framework for hyperspectral classification, where a lightweight student model is trained to mimic a larger teacher.
While effective, this approach typically depends on sufficiently annotated data to ensure reliable knowledge transfer.
Similarly, Cho et al.~\cite{cho2023pdp} introduced parameter pruning to compress neural models by removing redundant parameters.
Although such methods can reduce model size, aggressive compression may weaken spectral dependencies that are important for hyperspectral classification.
Therefore, these approaches do not directly address the joint challenge of communication efficiency, limited supervision, and spectral--spatial fidelity in satellite scenarios.

The third category is LoRA-based parameter-efficient fine-tuning, which has become a widely adopted paradigm for adapting pretrained models with a small number of trainable parameters.
LoRA represents weight updates using low-rank matrix pairs, thereby reducing computational, memory, and transmission overhead while preserving inference efficiency.
This framework has been successfully applied in natural language processing, computer vision, and multimodal learning \cite{mao2024doraenhancingparameterefficientfinetuning}.
Several variants have been proposed to further improve its effectiveness.
For instance, LoRA+ uses differentiated learning rates for low-rank factors to improve optimization in wide models \cite{hayou2024lora+}.
DoRA decomposes pretrained weights into magnitude and direction components and applies low-rank adaptation to the direction component to narrow the gap with full fine-tuning \cite{liu2024doraweightdecomposedlowrankadaptation}.
AdaLoRA adaptively allocates rank budgets across layers according to their importance, thereby improving parameter usage efficiency \cite{zhang2023adaloraadaptivebudgetallocation}.
Despite these advances, most existing LoRA variants still rely on linear low-rank update structures.
In hyperspectral applications, where inputs contain many spectral bands and exhibit complex spectral--spatial interactions, purely linear adapters may have limited capacity to capture the nonlinear patterns required for accurate adaptation.

Building on this analysis, we propose NE-LoRA, a dual-branch low-rank adaptation framework with differentiated training for multi-matrix adapters.
The proposed method introduces a nonlinear auxiliary branch to improve representational capacity and a differentiated training strategy to balance optimization across matrices, thereby achieving a better trade-off among adaptation accuracy, parameter efficiency, and transmission cost in hyperspectral satellite model updating.

\section{Design}
\subsection{Overview}\label{sec:subsection}
\begin{figure*}[htbp]
    \centering
    \includegraphics[width=\linewidth]{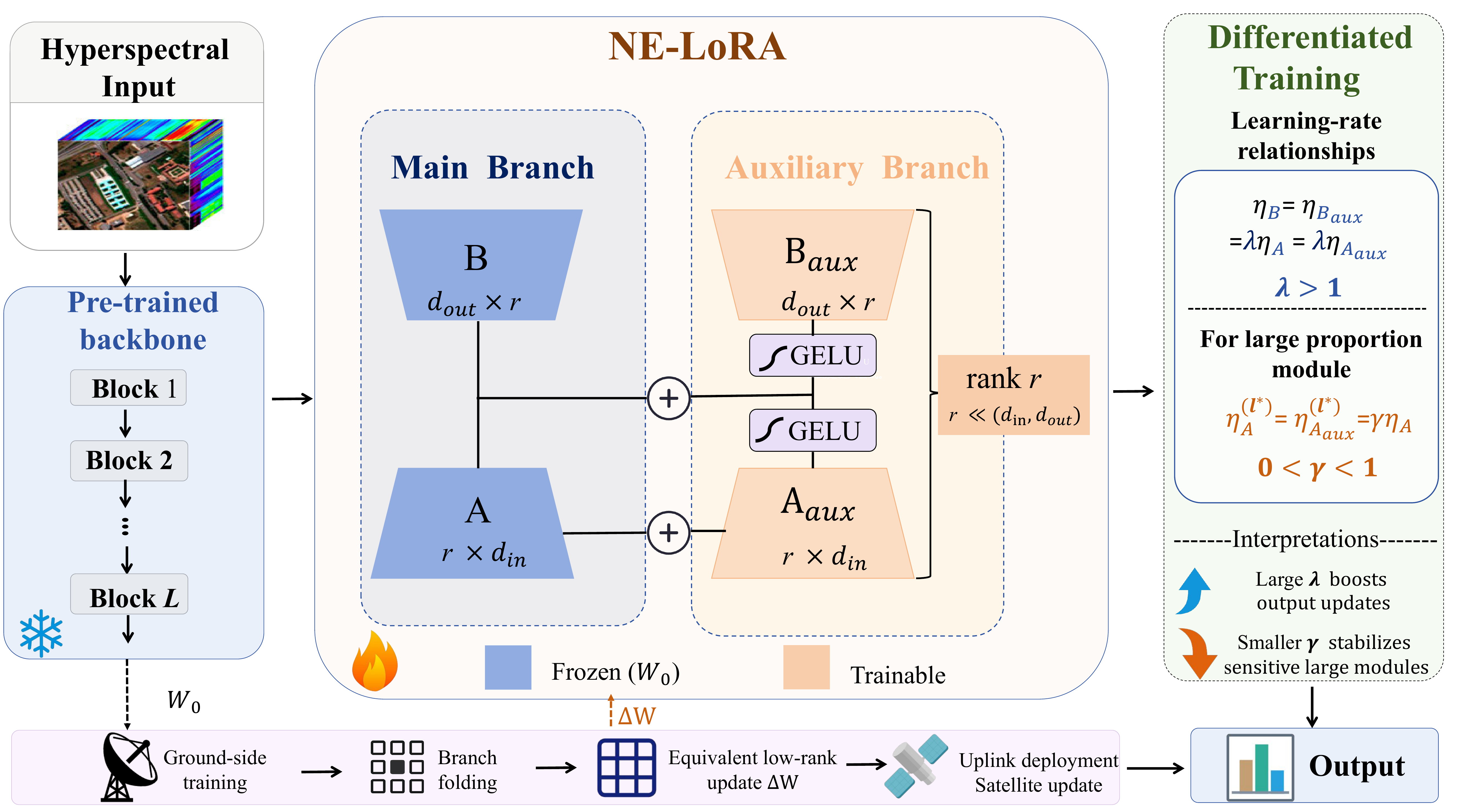}
    \caption{NE-LoRA overview.}
    \label{fig:over}
\end{figure*}

Fig.~\ref{fig:over} illustrates the overview design of NE-LoRA, which augments standard LoRA with nonlinear enhancement and differentiated training.
The framework adopts a dual branch design: a main low-rank branch that captures the dominant linear update with minimal trainable parameters, and an auxiliary branch activated by GELU to model complex spectral–spatial characteristics.
In addition, we introduce a differentiated training strategy for multi-matrix adapters, motivated by the asymmetric initialization and gradient dynamics of different adapter matrices in hyperspectral satellite models.
Together, these designs enable communication-efficient and stable adaptation, making the framework well suited for  bandwidth-constrained onboard hyperspectral applications.
During deployment, the auxiliary branch is folded into an equivalent low-rank update form used for transmission, so that the deployed upload cost remains the same order as standard LoRA. The branch merging is therefore a deployment-time reparameterization rather than transmission of a dense full-size update.

\subsection{Nonlinear-Enhanced Low-Rank Adaptation}\label{sec:1}
Parameter-efficient adaptation is a fundamental challenge for hyperspectral satellite models operating under strict uplink constraints.
Hyperspectral models joint process spectral and spatial information~\cite{li2019deep}; However, as the number of input channels increases, the parameter count grows rapidly. 
Each layer must handle high-dimensional spectral features,  often resulting in models with millions of parameters \cite{11007459}.
This not only increases the computational cost of fine-tuning, but also substantially raises the transmission cost of model updates, thereby limiting practical deployment in orbit.
   
To address the challenges, LoRA provides a natural starting point, as it reduces both computation and transmission overhead by representing weight updates with low-rank factors.
Instead of updating the full parameter matrix, LoRA optimizes only a small fraction of trainable parameters, making it particularly attractive for resource-constrained hyperspectral applications
Formally, for a model with pretrained weight $W_0 \in \mathbb{R}^{d \times k}$, where $d$ and $k$ denote the input and output dimensions, respectively, the adapted weight is written as:
\begin{equation}\label{eq:one}   
W = W_0 + \Delta W,
\end{equation}
where the weight increment $\Delta W = \alpha BA$, 
$B \in \mathbb{R}^{d \times r}$ and $A \in \mathbb{R}^{r \times k}$ with $r \ll \min\{d, k\}$, and $\alpha$ is a scaling factor that controls the update magnitude. 
$B$ is initialized with a Gaussian distribution, $A$ with zeros
Following the standard low-rank adaptation setup, one factor is randomly initialized while the other is initialized to zero so that the initial update remains close to zero.
Despite its efficiency, the strictly linear form of LoRA limits its ability to capture nonlinear spectral--spatial patterns, which are common in hyperspectral sensing~\cite{6555921}.

To improve representational capacity while preserving parameter efficiency, we propose NE-LoRA, 
a nonlinear-enhanced low-rank adaptation mechanism.
The key idea is to augment the standard low-rank update with an auxiliary nonlinear branch.
Specifically, the main branch ($A, B$) models the dominant linear update, while the auxiliary branch $(A_{\mathrm{aux}}, B_{\mathrm{aux}})$ captures additional nonlinear variations through GELU activation. 
Here, $A_{\mathrm{aux}} \in \mathbb{R}^{r \times k}$ and $B_{\mathrm{aux}} \in \mathbb{R}^{d \times r}$ have the same dimensionality as the corresponding matrices in the main branch.
Accordingly, the weight increment is defined as:
\begin{equation}
\Delta W
=
\alpha
\Big(
\big(B + \mathrm{GELU}(B_{\mathrm{aux}})\big)
\big(A + \mathrm{GELU}(A_{\mathrm{aux}})\big)
\Big),
\end{equation}
Only $A$, $B$, $A_{\mathrm{aux}}$, and $B_{\mathrm{aux}}$ are trainable, while $W_0$ remains frozen.
This design allows NE-LoRA to jointly model global linear trends and localized nonlinear corrections, both of which are important in heterogeneous and dynamic satellite environments.

The use of GELU is motivated by its smooth nonlinear behavior and favorable optimization properties \cite{DBLP:journals/corr/HendrycksG16}. In particular, GELU is approximately linear around small inputs while providing richer nonlinear responses for larger activations, which improves expressiveness without discarding the low-rank structure.
The GELU function is defined as:
\begin{equation}\label{equation3}
 \text{GELU}(x) = \frac{x}{2} \left(1 + \text{erf}\left( \frac{x}{\sqrt{2}} \right)\right)
\end{equation}
where $\text{erf}(\cdot)$ denotes the error function. 
Because the auxiliary matrices share the same dimensions as the main low-rank factors, their outputs remain naturally aligned with the main branch, which facilitates joint modeling of linear and nonlinear patterns while avoiding unnecessary parameter redundancy.

NE-LoRA also adopts a stability-aware initialization strategy. Specifically, the auxiliary matrices $A_{\mathrm{aux}}$ and $B_{\mathrm{aux}}$ and together with the main output-side matrix $B$, are initialized to zero, while the main input-side matrix $A$ is initialized with Kaiming initialization~\cite{He_2015_ICCV}.
At the beginning of training, since $A_{\mathrm{aux}} = B_{\mathrm{aux}} = B = \mathbf{0}$, the update is initially close to the standard LoRA form and avoids introducing large perturbations.
Notably, this asymmetric initialization directly leads to different gradient dynamics: the randomly initialized input-side matrix  inherently carries small perturbations whose gradients naturally weaken with model width, whereas the zero-initialized output-side matrices  and  lack effective gradient signals during early training and therefore require stronger updates to catch up.
This imbalance motivates the differentiated training strategy introduced in Section~\ref{sec:2}.
As training proceeds, the soft activation behavior of GELU gradually enables the auxiliary branch, yielding a smooth transition from purely linear adaptation to hybrid linear--nonlinear adaptation.
This progressive activation reduces optimization instability and improves training robustness.

\subsection{Differentiated Training for Accuracy Compensation}\label{sec:2}
Beyond representational capacity, adapting hyperspectral satellite models involves a second challenge: different adapter matrices contribute unevenly to optimization and exhibit asymmetric gradient dynamics.
%Standard LoRA applies a uniform learning rate to all low-rank factors, without distinguishing between input-side and output-side matrices.
Here, the input-side matrices project features into a low-rank subspace, whereas the output-side matrices map the adapted representation back to the original feature space.
This mismatch becomes even more pronounced in NE-LoRA, where multiple matrices are jointly optimized.

To address this issue, we introduce a differentiated training strategy tailored to NE-LoRA.
The strategy is motivated by two observations.
First, the adapter matrices in NE-LoRA are not optimization-wise equivalent.
The input-side and output-side matrices play different roles in the low-rank adaptation path, and their asymmetric initialization further leads to different gradient scales during training.
Consequently, applying a uniform learning rate to all matrices can disrupt the update balance, causing some matrices to accumulate unnecessary perturbations while others remain under-updated.
Second, this problem becomes more pronounced in large-proportion modules.
Since such modules contribute a substantial fraction of the overall parameter update, their optimization dynamics exert a disproportionately strong influence on the whole adaptation process.
If their learning rates are not properly controlled, the resulting updates can dominate the training trajectory and degrade training stability.
Accordingly, our strategy includes two components: (i) learning-rate differentiation across adapter matrices to restore update balance, and (ii) learning-rate reduction for the identified large-proportion module to improve training stability. 
%An overview of this design is provided in Fig.~\ref{fig:model}.
%Fig.~\ref{fig:model} clearly presents the key difference between standard LoRA and our proposed NE-LoRA in terms of learning rate setting: standard LoRA uses the same learning rate for matrices $A$ and $B$, failing to account for their asymmetric initialization and gradient dynamics and thus leading to suboptimal fine-tuning; in contrast, NE-LoRA adopts a tailored configuration—setting the learning rate of $B_1, B_2$ to $\lambda \times$ that of $A_1, A_2$ (with $\lambda \gg 1$ as a fixed value), and for large-proportion modules, setting the learning rate of $A_1^{\text{large}}, A_2^{\text{large}}$ to $\gamma \times$ that of $B_1^{\text{large}}, B_2^{\text{large}}$ (with $\gamma \ll 1$ as a fixed value).

We first address the imbalance caused by the asymmetric initialization and gradient dynamics of multi-matrix structure. 
In particular, the input-side matrices $(A, A_{\mathrm{aux}})$ and output-side matrices $(B, B_{\mathrm{aux}})$ play different roles during adaptation and should therefore not share the same learning rate.
The matrix $A$ is initialized with Kaiming initialization to prevent excessive feature scaling, while $A_{\mathrm{aux}}$, $B$, and $B_{\mathrm{aux}}$ are initialized to zero to ensure stable adaptation from pretrained weights.
As model width increases, this initialization asymmetry leads to different gradient scales: the randomly initialized input-side matrices can introduce small perturbations from the beginning of training, whereas the zero-initialized output-side matrices tend to receive weaker effective updates in the early stage.
% \textcolor{blue}{$A$, with its random initialization, inherently carries small perturbations whose gradients weaken with width, in contrast, $B$ and $B_1$, initialized at zero, lack gradient signals during early training and require stronger updates.}
Using a uniform learning rate therefore either over-amplifies perturbations in the input-side matrices or under-updates the output-side matrices, resulting in suboptimal adaptation.

%\textcolor{red}{Change2(lxh):We refer to LoRA+ for λ; LoRA+ also introduces this factor. We add an extra branch and apply this factor to both branches. Do we need to emphasize "inspired by LoRA+"?}

To mitigate this issue, we assign a larger learning rate to the output-side matrices.
Specifically, we set the learning rates of $B$ and $B_{\mathrm{aux}}$ to $\lambda$ times those of $A$ and $A_{\mathrm{aux}}$, where $\lambda > 1$:
$
\eta_B = \eta_{B_{\mathrm{aux}}} = \lambda \eta_A = \lambda \eta_{A_{\mathrm{aux}}}.
$
This design improves optimization in three ways: First, a larger learning rate compensates for the weak early-stage gradients of $B$ and $B_{\mathrm{aux}}$ caused by zero initialization. 
Second, a smaller learning rate for $A$ and $A_{\mathrm{aux}}$ suppresses the accumulation of unnecessary perturbations and stabilizes training.
Third, the fixed ratio $\lambda$ maintains a balanced update relationship between input-side and output-side matrices as model width varies.
Therefore, $\lambda > 1$ serves as a principled training modulation factor rather than an arbitrary amplification coefficient.
%\textcolor{red}{At training onset, $A$ achieves stable updates, while $A_1$, $B$, and $B_1$ exhibit little effective change. Especially $B$, whose initial update magnitude is nearly zero. Scaling factor $\lambda$ accelerates $B$/$B_1$’s learning, gradually aligning their update magnitudes with $A$’s, thereby balancing all matrices’ contributions to fine-tuning.}

After addressing the imbalance across adapter matrices, we further consider instability caused by large-proportion modules.
Such modules account for a substantial fraction of the parameter update and are particularly sensitive to learning-rate choices, as illustrated in Appendix.
Prior research overlooked the low-rank adaptation characteristics, despite their large parameter scales and strong influence on overall performance \cite{zeng2024expressivepowerlowrankadaptation}. 
Without tailored control, they may introduce excessively large or insufficient updates, thereby destabilizing the overall adaptation process.
To identify such modules, we denote by $\Delta W^{(l)}$ the update of the $l$-th adapted layer and define
\begin{equation}
l^\star = \arg\max_l \|\Delta W^{(l)}\|_F ,
\end{equation}
where $\|\cdot\|_F$ is the Frobenius norm.
The layer indexed by $l^\star$ is treated as the large-proportion module.

For this modules, we introduce a reduction factor $\gamma$ ($0 < \gamma < 1$), to further modulate the learning rates of its input-side matrices:
\begin{equation}
\eta_A^{(l^\star)} = \eta_{A_{\mathrm{aux}}}^{(l^\star)} = \gamma \eta,
\qquad
\eta_B^{(l^\star)} = \eta_{B_{\mathrm{aux}}}^{(l^\star)} = \eta.
\end{equation}
%The input-side learning rate is set to $\eta_{A,\text{large}} = \gamma \cdot \eta$, where $\eta$ is the base learning rate; while the output-side learning rate is fixed as $\eta_{B,\text{large}} = \eta$. 
%Consequently, the learning rate ratio between them is $\eta_{A,\text{large}} / \eta_{B,\text{large}} = \gamma$, directly determined by the reduction factor.
This design extends the logic of differentiated training by applying additional control to the most sensitive module.
Reducing the learning rate of its input-side matrices alleviates the risk of overly aggressive updates caused by large dimensionality, while preserving sufficient adaptation capacity on the output side.

%Overall, the proposed differentiated training strategy improves optimization stability from two complementary perspectives: it compensates for the asymmetric initialization and gradient dynamics across adapter matrices, and it regularizes updates in large-proportion modules that are particularly sensitive to learning-rate selection.

\section{Implementation and Methodology}

\textbf{Implementation Details.}
Our method is implemented in Python 3.8 with PyTorch 2.2 and CUDA 12.1; all experiments are conducted on NVIDIA RTX 4090 GPUs. 
For hyperparameter exploration, we vary the rank $r$ over $\{1, 3, 5, 7, 9, 11, 13, 15\}$ to examine the effect of different low-rank constraints.
In particular, we also evaluate the extreme low-rank case $r=1$ to assess whether NE-LoRA can still capture high-level hyperspectral semantics under a highly constrained adaptation budget.
We further vary the learning-rate ratio $\lambda$ over $\{0.0001, 0.0005, 0.001, 0.005, 0.01, 0.05, 0.1, 0.2, 0.3, 0.4, 0.5, 0.6, 0.7, 0.8, 0.9, 1, 2, 3, 4, 5, 6, 7, 8, 9, 10\}$ to study its effect on differentiated optimization.
%Overall, the method does not rely on heavy hyperparameter tuning: extensive experiments show that both $\lambda$ ($\lambda > 1$) and $\gamma$ ($0<\gamma<1$) consistently improve performance when chosen within reasonable non-extreme ranges.
In practice, we observe that $\lambda \in [2,8]$ and $\gamma \in [0.2,0.8]$ provide stable and effective performance across datasets.

\textbf{Models.} We evaluate three representative backbone models widely used in hyperspectral image analysis: CNN3D \cite{DBLP:journals/corr/abs-1806-05824}, which directly captures joint spectral--spatial features through 3D convolutions; M3DDCNN \cite{8297014}, a multi-scale 3D convolutional neural network with hierarchical feature fusion for complex spectral variations; and HybridSN \cite{Roy_2020}, a hybrid network combining 3D convolutions and 2D convolutions for hyperspectral image classification.
Although LoRA is more commonly studied in large foundation models, our choice of lightweight backbones is motivated by the computational constraints of on-orbit platforms.
Even for compact models, parameter-efficient adaptation remains important because full-parameter transmission still incurs prohibitive uplink cost in orbit.

\textbf{Datasets.} We conduct experiments on four hyperspectral image datasets: (i) Salinas Valley (SA) \cite{deng2019pesticide}: with 204 bands and 16 classes; (ii) Pavia University (PU) \cite{kavitha2014classification}: with 103 bands and 9 classes; (iii) HyRANK-Loukia (HR-L) \cite{karantzalos2018hyrank}: with 176 bands, spatial resolution $249 \times 945$, and 14 classes; (iv) WHU-LongKou (WHU-LK) \cite{ZHONG2020112012}: with 270 bands and 9 classes. For data partitioning, we use 1\% of samples for initial training and 1\% for LoRA updates on SA and PU, 0.5\% + 0.5\% on WHU-LK, and 3\% + 3\% on HR-L.
These settings are consistent with common practice in hyperspectral classification, where limited labeled samples are often sufficient for effective adaptation.
%Together, these datasets cover diverse spectral characteristics and scene complexities, providing a broad evaluation of model robustness and generality.

\textbf{Metrics.} We adopt three classification metrics to evaluate model performance \cite{grandini2020metricsmulticlassclassificationoverview}: 
(i) \textit{Overall Accuracy} (OA), which measures the proportion of correctly classified pixels;
(ii) \textit{Kappa coefficient}, which quantifies agreement beyond chance and complements OA in assessing prediction reliability; and
(iii) \textit{Average Accuracy} (AA), which averages class-wise accuracies and reflects performance across different categories. 
To evaluate parameter efficiency, we additionally report three parameter-related metrics: Total parameters, the total number of model parameters; Trainable parameter, the number of parameters optimized during ground-side training; Upload parameters, the number of parameters that must be transmitted to the satellite during deployment. 

\textbf{Stability-Oriented Configuration.} Unless otherwise specified, we employ consistent settings across all methods to ensure a fair comparison. We evaluate the framework's sensitivity to the rank
$r$, the learning rate ratio $\lambda$ for output-side matrices, and the reduction factor for large modules. Rather than relying on exhaustive hyper-parameter tuning, our objective is to demonstrate that NE-LoRA provides stable and effective performance across non-extreme ranges, which is critical for autonomous in-orbit operation. The evaluated ranges for these factors are detailed in Appendix A.4.

\textbf{Baselines.} We compare NE-LoRA against five baselines: (i) No Fine-tuning (NFT), which performs no parameter updates; (ii) Full Fine-tuning (FT), which updates all model parameters;  (iii) LoRA \cite{hu2022lora}, which freezes pretrained weights and introduces trainable low-rank adapters for downstream adaptation; (iv) LoRA+ \cite{hayou2024lora+}, which assigns different learning rates to LoRA’s adapter to improve optimization in wide models; and (v) VERA \cite{kopiczko2024veravectorbasedrandommatrix}, which shares a single pair of low-rank matrices across layers and learns lightweight scaling vectors to reduce the number of trainable parameters.

\begin{table}[t]
\centering
\setlength{\tabcolsep}{2.5pt}
\renewcommand{\arraystretch}{0.8}
\scriptsize
\caption{Performance of different methods on CNN3D.
Numbers in parentheses indicate the average improvement of NE-LoRA over the compared methods for each column. The numbers in parentheses indicate the improved precision or the reduction in uplink parameters compared to the FT.}
\label{tab:cnn3d_performance}
\begin{tabular}{
>{\centering\arraybackslash}m{0.06\linewidth}
>{\centering\arraybackslash}m{0.11\linewidth}
>{\centering\arraybackslash}m{0.11\linewidth}
>{\centering\arraybackslash}m{0.11\linewidth}
>{\centering\arraybackslash}m{0.12\linewidth}
>{\centering\arraybackslash}m{0.08\linewidth}
>{\centering\arraybackslash}m{0.08\linewidth}
>{\centering\arraybackslash}m{0.12\linewidth}
}
\toprule
\textbf{Dataset} & \textbf{Method} & \textbf{OA (\%)} & \textbf{AA (\%)} & \textbf{Kappa ($\times 100$)}
& \textbf{Total} & \textbf{Train} & \textbf{Upload} \\
\midrule

\multirow{6}{*}{SA}
& NFT& 91.98 & 94.22 & 91.07 & 161.7k & 0 & 0 \\
& FT & 93.22 & 95.67 & 92.45 & 161.7k & 161.7k & 161.7k \\
& LoRA & 93.13 & 95.71 & 92.35 & 189.4k & 27.7k & 27.7k \\
& LoRA+ & 93.15 & 96.12 & 92.39 & 189.4k & 27.7k & 27.7k \\
& VERA & 90.55 & 93.79 & 89.47 & 190.0k & 0.7k& 0.7k\\
& \textbf{NE-LoRA}
& \textbf{94.37 (+1.96)}
& \textbf{96.78 (+1.68)}
& \textbf{93.74 (+2.19)}
& \textbf{217.1k} & \textbf{55.4k} & \textbf{27.7k (5.8$\times$)} \\
\midrule

\multirow{6}{*}{PU}
& NFT& 93.50 & 90.96 & 91.28 & 70.3k & 0 & 0 \\
& FT & 96.90 & 95.44 & 95.90 & 70.3k & 70.3k & 70.3k \\
& LoRA & 96.48 & 94.55 & 95.34 & 86.6k & 16.3k & 16.3k \\
& LoRA+ & 96.75 & 95.16 & 95.70 & 86.6k & 16.3k & 16.3k \\
& VERA & 90.19 & 82.74 & 86.78 & 87.3k & 0.7k & 0.7k \\
& \textbf{NE-LoRA}
& \textbf{97.42 (+2.66)}
& \textbf{96.30 (+4.53)}
& \textbf{96.59 (+3.59)}
& \textbf{103.0k} & \textbf{32.7k} & \textbf{16.3k (4.3$\times$)} \\
\midrule

\multirow{6}{*}{HR-L}
& NFT & 78.95 & 71.01 & 74.92 & 132.1k & 0 & 0 \\
& FT & 81.89 & 77.91 & 78.49 & 132.1k & 132.1k & 132.1k \\
& LoRA & 84.93 & 79.19 & 81.96 & 156.9k & 24.8k & 24.8k \\
& LoRA+ & 84.20 & 78.01 & 80.96 & 156.9k & 24.8k & 24.8k \\
& VERA & 77.53 & 65.44 & 72.89 & 157.6k & 0.7k & 0.7k \\
& \textbf{NE-LoRA}
& \textbf{87.43 (+5.93)}
& \textbf{84.23 (+9.92)}
& \textbf{85.03 (+7.19)}
& \textbf{181.8k} & \textbf{49.7k} & \textbf{24.8k (5.3$\times$)} \\
\midrule

\multirow{6}{*}{WHU-LK}
& NFT& 84.19 & 50.88 & 78.87 & 127.0k & 0 & 0 \\
& FT & 87.78 & 55.88 & 83.63 & 127.0k & 127.0k & 127.0k \\
& LoRA & 88.56 & 58.61 & 84.75 & 162.2k & 35.2k & 35.2k \\
& LoRA+ & 89.95 & 65.99 & 86.71 & 162.2k & 35.2k & 35.2k \\
& VERA & 80.94 & 47.38 & 74.45 & 162.9k & 0.7k & 0.7k \\
& \textbf{NE-LoRA}
& \textbf{90.29 (+4.01)}
& \textbf{67.63 (+11.88)}
& \textbf{87.15 (+5.47)}
& \textbf{197.5k} & \textbf{70.5k} & \textbf{35.2k (3.6$\times$)} \\
\bottomrule
\end{tabular}
\end{table}

\section{Evaluation}
\subsection{Performance Analysis}
%Tables \ref{tab:cnn3d_performance}, \ref{tab:m3cnn3d_performance}, and \ref{tab:hybridsn_performance} present the accuracy and parameter counts of the baseline methods in three models and four datasets. These results confirm the superiority of NE-LoRA over baseline methods, and consistently achieves a balance between accuracy and parameter efficiency, while effectively adapting to diverse models and datasets. 
%Table \ref{tab:cnn3d_performance} summarizes the classification performance and parameter cost of all methods. NE-LoRA consistently improves over LoRA-based baselines across different datasets and backbone architectures, while remaining competitive with, and in several cases superior to, full fine-tuning. These results suggest that adding nonlinear refinement and differentiated training can substantially improve the adaptation capability of low-rank updates for hyperspectral data.
%The convergence of all NE-LoRA data points in the top-left corner of Figure X confirms its effectiveness for bandwidth-constrained satellite intelligence, providing a scalable solution for high-fidelity in-orbit model evolution.

\begin{figure}[t]
  \centering
  % 左侧文字部分
  \begin{minipage}{0.48\textwidth}
    \vspace{0pt} % 确保顶端对齐
Table \ref{tab:cnn3d_performance} summarizes the classification performance and parameter cost of all methods. NE-LoRA consistently achieves a decisive "low-cost, high-performance" advantage, yielding an average 3.2\% OA gain over FT. TAs visually demonstrated by the convergence of all NE-LoRA data points in the top-left corner of Fig. \ref{fig:tradeoff}, our framework provides a highly scalable solution for high-fidelity onboard model updates under severe bandwidth constraints. These results suggest that integrating nonlinear refinement with a differentiated training strategy substantially improves the adaptation capability of low-rank updates for complex hyperspectral data.
  \end{minipage}
  \hfill % 在两个 minipage 之间填充空白
  % 右侧图片部分
  \begin{minipage}{0.48\textwidth}
    \vspace{0pt} % 确保顶端对齐
    \centering
    \includegraphics[width=\textwidth]{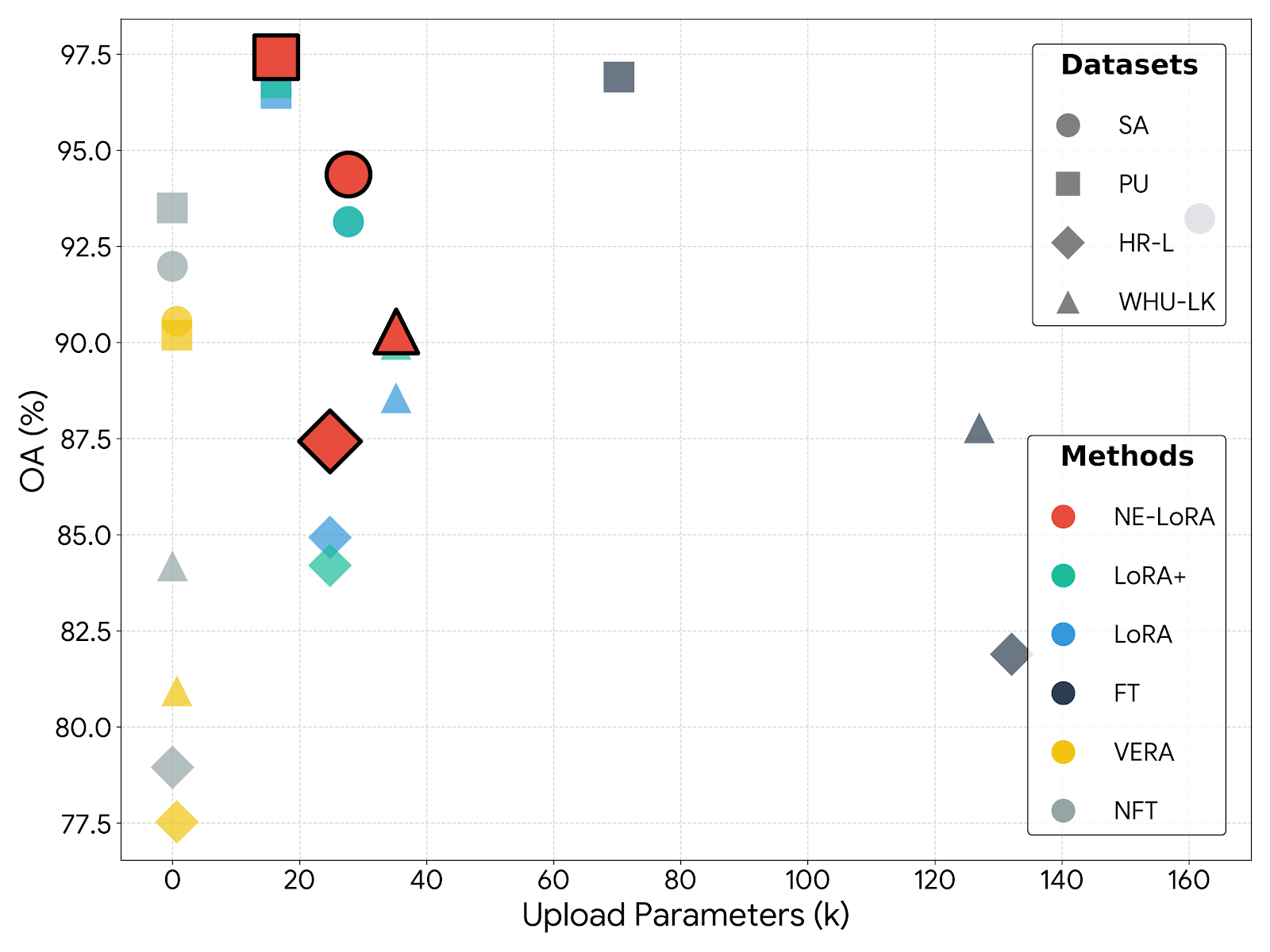}
    \caption{Accuracy vs. Communication.}
    \label{fig:tradeoff}
  \end{minipage}
\end{figure}

\textbf{Classification Accuracy.} 
%This metric serves as a fundamental measure for evaluating the adaptation performance of large models in hyperspectral satellite applications. From the Table \ref{tab:cnn3d_performance}, it is evident that NE-LoRA not only outperforms LoRA and its variants across various metrics, but also surpasses FT in the majority of cases. Specifically, on CNN3D, FT achieves the highest accuracy by updating all weights and fully capturing spectral–spatial correlations. NFT suffers substantial degradation due to fixed pre-trained weights, with OA consistently 1–7\% lower than FT.  LoRA and LoRA+ improve upon NFT but exhibit limited adaptability and unstable performance: both approach FT on the SA dataset, but on HR-L, LoRA outperforms NFT while LoRA+ drops to 84.20\%, indicating inconsistent adaptation to spectral-overlapping scenes. NE-LoRA consistently outperforms LoRA and LoRA+ on CNN3D across all datasets, while matching or surpassing the FT baseline. This is attributed to its dual-branch structure, which better models complex spectral-spatial features via GELU-activated nonlinear enhancement.
Classification accuracy is the primary metric for assessing adaptation quality in onboard hyperspectral applications. Compared with NFT, all adaptation methods generally improve performance, confirming the necessity of updating pre-trained weights for evolving data distributions. NE-LoRA delivers more consistent gains across datasets than LoRA and LoRA+, particularly on challenging scenes with strong spectral overlap. For instance, on HR-L, NE-LoRA achieves an OA of 87.43\%, representing a significant 5.93\% improvement over the standard LoRA. In several settings, NE-LoRA also matches or exceeds full fine-tuning, indicating that a carefully designed parameter-efficient update can be more effective than naively updating all parameters. On PU, NE-LoRA reaches 97.42\% OA, surpassing the 96.90\% achieved by FT.

\textbf{Parameter Efficiency.} 
Communication efficiency is a first-class requirement in satellite deployment due to the stringent uplink bottlenecks. NE-LoRA updates only 18\% of total parameters on average while maintaining or even exceeding FT-level accuracy. On WHU-LK, NE-LoRA reduces the number of upload parameters from 127.0k (FT) to just 35.2k, an approximately 72\% reduction in communication volume without sacrificing performance. 
By leveraging a branch-folding deployment strategy, NE-LoRA maintains a lightweight upload footprint identical to standard LoRA while providing significantly higher representational capacity.
Under the real-world 12 Mbps uplink constraint, NE-LoRA reduces the transmission delay from 431.2 ms to 73.9 ms on SA, achieving a 5.8$\times$ speedup over FT. This efficiency ensures that model updates can be reliably completed within short satellite-ground contact windows, even as model complexity grows.
%This is a key factor that affects the training cost and deployment feasibility of large hyperspectral satellite models. Given the strict constraints on parameter transmission and storage in satellite environments,  on average, NE-LoRA updates merely 18\% of the total parameters, yet it achieves 3.2\% higher accuracy than models trained with full fine-tuning, shows its efficiency and effectiveness. FT requires full-model weight updates, resulting in consistently high total, trainable, and upload parameter counts across all cases. NE-LoRA adds extra low-rank matrices, slightly increasing the total parameter count compared to FT, but drastically reducing trainable and upload parameters, which are identical, thereby lowering both training and deployment costs.  Specifically, across the four datasets, these upload parameters average 21.7\% of FT's on CNN3D. NE-LoRA introduces more trainable parameters than LoRA, but this increase is intentional: the added lightweight refinement branch complements LoRA's low-rank matrices by enhancing fine-grained feature modeling and preventing bottlenecks from insufficient adjustable parameters.

\begin{table}[t]
\centering
\renewcommand{\arraystretch}{0.85}
\footnotesize
\caption{Ablation analysis of different methods for each model.}
\label{tab:4method_model_performance}
\begin{tabular}{
>{\centering\arraybackslash}m{0.15\linewidth}
>{\centering\arraybackslash}m{0.23\linewidth}
>{\centering\arraybackslash}m{0.10\linewidth}
>{\centering\arraybackslash}m{0.10\linewidth}
>{\centering\arraybackslash}m{0.10\linewidth}
>{\centering\arraybackslash}m{0.15\linewidth}
}
\toprule
\textbf{Model} & \textbf{Method} & \textbf{SA (\%)} & \textbf{PU (\%)} & \textbf{HR-L (\%)} & \textbf{WHU-LK (\%)} \\
\midrule

\multirow{4}{*}{CNN3D}
& GELU + $\lambda$ + $\gamma$ & \textbf{94.37} & \textbf{97.42} & \textbf{87.43} & 90.29 \\
& GELU + $\lambda$            & 94.06          & 96.97          & 85.97          & \textbf{90.71} \\
& $\lambda + \gamma$          & 93.73          & 97.01          & 84.98          & 89.23 \\
& GELU                        & 92.37          & 96.68          & 84.16          & 88.78 \\
\midrule

\multirow{4}{*}{M3DDCNN}
& GELU + $\lambda$ + $\gamma$ & \textbf{94.35} & \textbf{97.06} & \textbf{84.57} & \textbf{91.85} \\
& GELU + $\lambda$            & 93.75          & 96.68          & 83.06          & 91.00 \\
& $\lambda + \gamma$          & 93.90          & 96.00          & 83.73          & 91.47 \\
& GELU                        & 92.85          & 96.02          & 82.87          & 91.71 \\
\midrule

\multirow{4}{*}{HybridSN}
& GELU + $\lambda$ + $\gamma$ & 90.84          & \textbf{97.69} & \textbf{84.76} & \textbf{89.98} \\
& GELU + $\lambda$            & \textbf{91.82} & 97.07          & 84.68          & 88.22 \\
& $\lambda + \gamma$          & 89.57          & 96.98          & 84.24          & 84.81 \\
& GELU                        & 88.76          & 96.99          & 84.67          & 89.16 \\
\bottomrule
\end{tabular}
\end{table}
\subsection{Ablation Studies}
Table \ref{tab:4method_model_performance} presents the ablation results and highlights the complementary roles of the three key components in NE-LoRA.
The GELU branch improves representational capacity, $\lambda$ restores update balance across adapter matrices, and $\gamma$ enhances training stability in large-proportion modules.
%Across different datasets and backbone models, 
Introducing the nonlinear GELU branch consistently improves classification accuracy over GELU-free variants.
This result suggests that the auxiliary nonlinear branch effectively compensates for the limited expressiveness of purely linear low-rank updates in complex spectral--spatial nonlinearities of hyperspectral data.
Incorporating $\lambda$ further improves performance by compensating for the asymmetry between input matrices and output matrices, ensuring balanced feature updates.
The reduction factor $\gamma$ provides an additional gain in settings that contain large-proportion modules, particularly for HybridSN on the HR-L dataset.
%Overall, the ablation results verify that the three components are complementary.

%Their combination yields the most consistent accuracy gains across diverse hyperspectral datasets and model architectures.

%\input{Related}
\section{Conclusion and Discussion}

This work proposes NE-LoRA, a nonlinear-enhanced and differentiated adaptation framework for hyperspectral satellite models.
NE-LoRA combines a GELU-activated auxiliary branch with customized learning-rate strategies designed to address asymmetric optimization dynamics across adapter matrices and instability in large-proportion modules.
%Experiments on multiple hyperspectral datasets and backbone architectures demonstrate consistent improvements over LoRA-based baselines.
%Across the evaluated settings, NE-LoRA updates only 18\% of the total parameters on average while achieving an average OA gain of 3.2\% over full fine-tuning, indicating a favorable accuracy--communication trade-off for bandwidth-constrained onboard adaptation.
NE-LORA delivers a decisive "low-cost, high-performance" advantage by achieving state-of-the-art classification accuracy while updating only a small fraction of total parameters, proving that high-fidelity hyperspectral model evolution is feasible even under the most stringent satellite uplink bandwidth constraints.

Our work also has several limitations.
The current evaluation is centered on hyperspectral classification models and does not yet cover larger or more general pretrained architectures, which limits the present evidence for broader cross-domain generalization.
Moreover, we focus on a practically motivated ground-side training and onboard deployment pipeline, but do not explicitly study long-term continual updating under time-varying orbital connectivity and resource dynamics.
Finally, while NE-LoRA reduces the parameter transmission burden, a tighter integration with system-level factors such as contact-window scheduling, update latency, and multi-satellite coordination remains an important direction for future work.

\bibliographystyle{plainnat}
\bibliography{sample-base}

%%%%%%%%%%%%%%%%%%%%%%%%%%%%%%%%%%%%%%%%%%%%%%%%%%%%%%%%%%%%

\appendix

\section{Appendix}
\subsection{Hyperspectral Dataset Overview}
\begin{figure}[!t]
   %\captionsetup{skip=-2pt}
    \centering
    \begin{minipage}[b]{0.47\linewidth}
        \centering
        % \vspace{0cm}  
        \includegraphics[width=\linewidth]{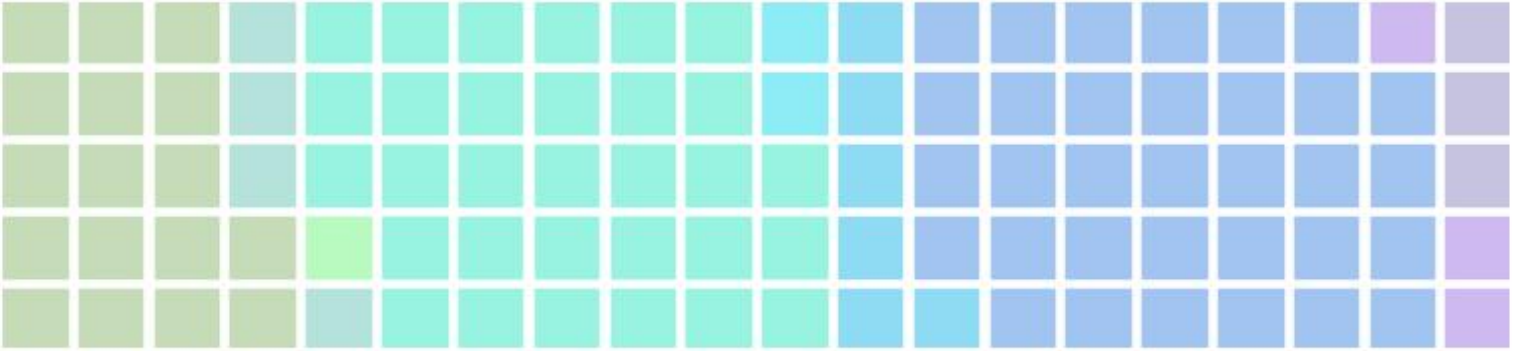}
        % \vspace{-0.1cm}  
        % 关键修改：\small → \footnotesize
        \scriptsize\textbf{\hspace{0.2cm}WHU-LK}  
        \label{fig:sub1}
    \end{minipage}
    \hspace{0.02\linewidth}  
    \begin{minipage}[b]{0.47\linewidth}
        \centering
        % \vspace{0.2cm}  
        \includegraphics[width=\linewidth]{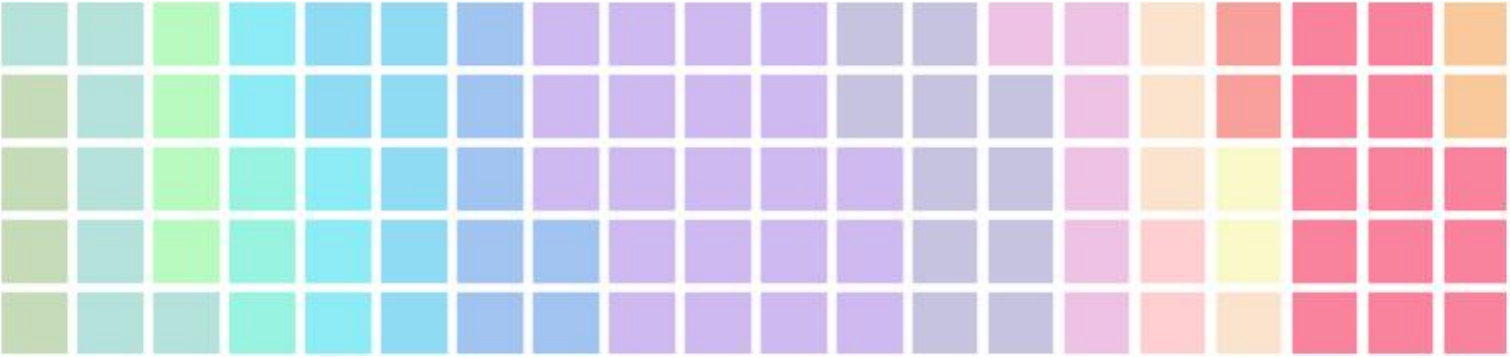}
        % \vspace{-0.1cm}  
        % 关键修改：\small → \footnotesize
        \scriptsize\textbf{\hspace{0.2cm}SA}  
        \label{fig:sub2}
    \end{minipage}
    \begin{minipage}[b]{0.47\linewidth}
        \centering
        % \vspace{0.2cm}  
        \includegraphics[width=\linewidth]{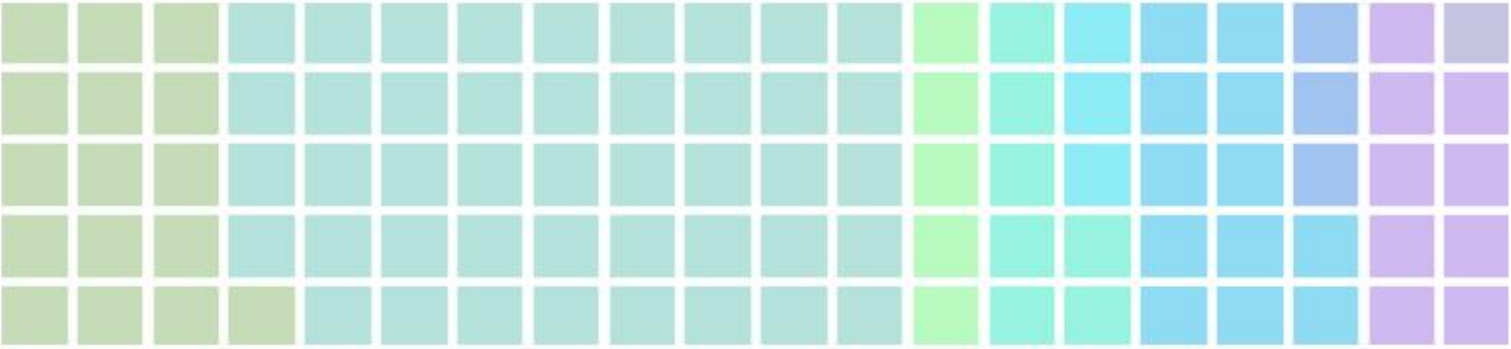}
        % \vspace{-0.1cm}  
        % 关键修改：\small → \footnotesize
        \scriptsize\textbf{\hspace{0.2cm}PU}  
        \label{fig:sub3}
    \end{minipage}
    \hspace{0.02\linewidth}  
    \begin{minipage}[b]{0.47\linewidth}
        \centering
        % \vspace{0.2cm}  
        \includegraphics[width=\linewidth]{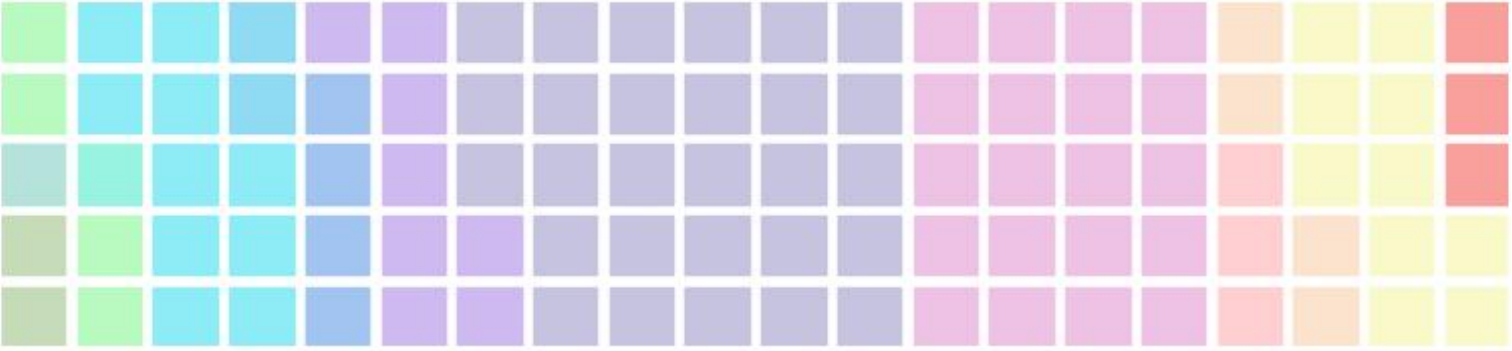}
        % \vspace{-0.1cm}  
        % 关键修改：\small → \footnotesize
        \scriptsize\textbf{\hspace{0.2cm}HR-L}  
        \label{fig:sub4}
    \end{minipage}

    \caption{\small Hyperspectral dataset overview. Different colors in the upper cells denote various ground object categories within each dataset, where color proportion indicates the relative prevalence of each class.}
    \vspace{-5pt}
    \label{fig:3}
\end{figure}

Fig.~\ref{fig:3} illustrates the category distributions of the four hyperspectral datasets.
The panels visualize the class composition of each dataset, highlighting their diversity in both category coverage and class proportion.
This diversity indicates that the evaluated datasets span heterogeneous scene characteristics, making them suitable for assessing the robustness and generality of the proposed method.

\subsection{Distribution of Parameter Proportions}
\begin{figure}    
\centering    
\includegraphics[width=0.6\linewidth]{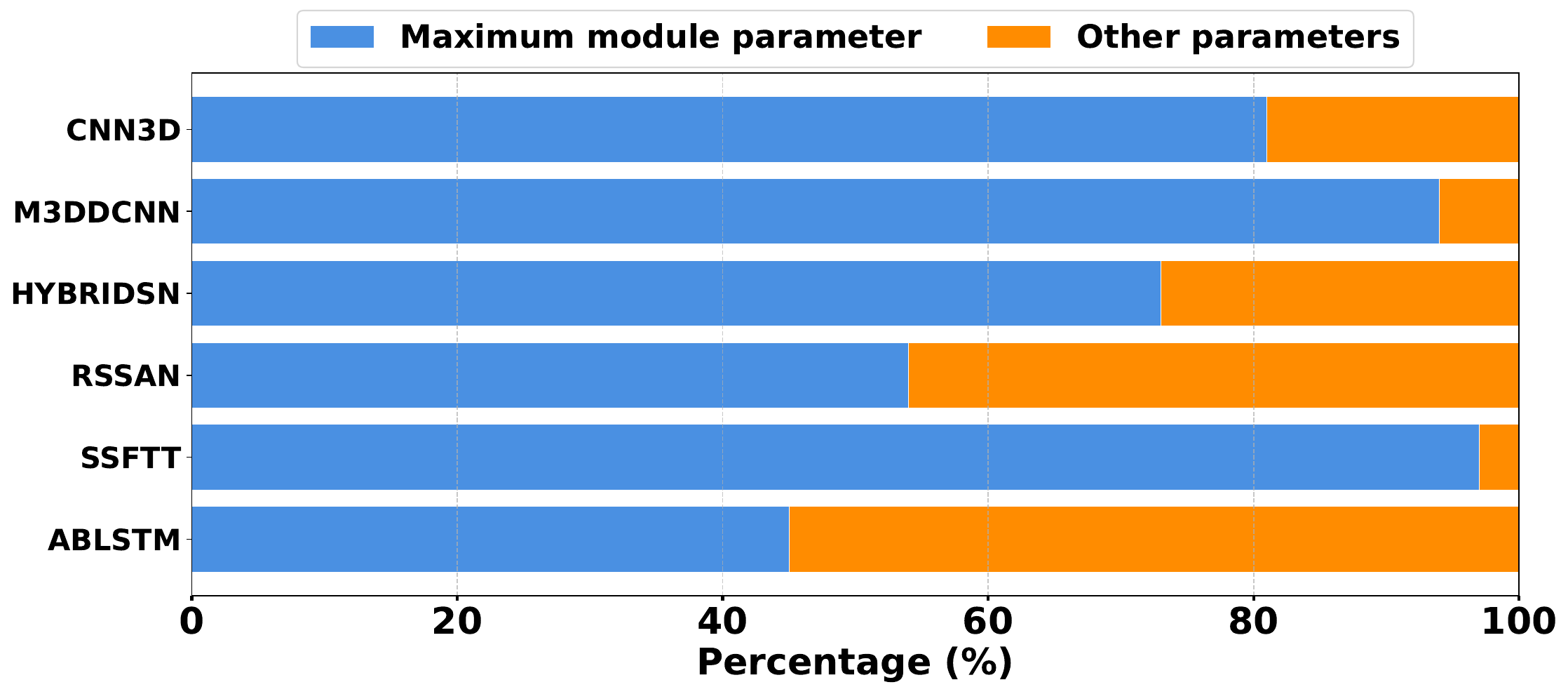}    
\caption{Distribution of parameter proportions for maximum-volume modules across different models.}    
\label{fig:777}
\vspace{-0.5em}
\end{figure}
Fig.~\ref{fig:777} shows the parameter proportion of the largest module in each backbone model.
The results indicate that several representative hyperspectral backbones contain one dominant module that accounts for a substantial fraction of the total parameters.
This observation motivates the large-proportion module analysis in our differentiated training strategy.
Because such modules contribute disproportionately to the overall update, they are more sensitive to learning-rate choices and can strongly influence optimization stability.

\subsection{Diverse Datasets}
On M3DDCNN, FT a strong high-accuracy baseline and achieves the best OA in several settings.
LoRA and LoRA+ perform similarly on PU but generally do not match FT.
NE-LoRA maintains clear advantages on M3DDCNN, particularly on SA and HR-L, where it adapts effectively to scene characteristics, outperforming LoRA and LoRA+ while approaching or matching FT. These results suggest that the proposed differentiated training strategy improves optimization under asymmetric gradient dynamics and supports stable adaptation across diverse scenes.

On HybridSN, FT again provides a strong baseline, whereas NFT and VERA perform relatively poorly. 
LoRA and LoRA+ yield only modest gains, with LoRA+ narrowing but not closing the gap with FT.
By contrast, NE-LoRA achieves the best or near-best performance on several datasets: it surpasses FT on PU (OA: 97.43\%) and WHU-LK (OA: 89.74\%), ranks second only to FT on SA (OA: 90.84\%), and remains competitive on HR-L. These results indicate that combining nonlinear enhancement with differentiated training can effectively compensate for the limitations of purely linear low-rank adaptation.
Although the absolute OA gains of NE-LoRA are sometimes modest, typically around 1--5\%, such improvements are meaningful for hyperspectral image classification, where spectral--spatial distinctions are often subtle and difficult to model.
In particular, HR-L yields the lowest accuracies across most methods because of its heterogeneous land-cover patterns and spectral ambiguity.
Even in this challenging setting, NE-LoRA still demonstrates robust fine-tuning capability, suggesting that proposed method generalizes well to complex hyperspectral scenes.

Across the four datasets, these upload parameters average 30.6\% on M3DDCNN, and just 2.5\% on HybridSN.
Crucially, NE-LoRA maintains the same upload parameter count as LoRA through its branch-folding deployment strategy.
The auxiliary refinement branch improves training but is folded into the deployed update afterward, so only lightweight low-rank parameters need to be transmitted.
By contrast, although VERA updates very few parameters, its post-update accuracy remains unsatisfactory.
This result suggests that hyperspectral data, characterized by complex spectral overlap and spatial heterogeneity, cannot always be adapted effectively with extremely small update budgets.

\begin{table}[t]
\centering
\setlength{\tabcolsep}{2.5pt}
\renewcommand{\arraystretch}{0.9}
\scriptsize
\caption{Performance of different methods on M3DDCNN.
Numbers in parentheses indicate the average improvement of NE-LoRA over the compared methods for each column.}
\label{tab:m3ddcnn_performance}
\begin{tabular}{
>{\centering\arraybackslash}m{0.06\linewidth}
>{\centering\arraybackslash}m{0.11\linewidth}
>{\centering\arraybackslash}m{0.11\linewidth}
>{\centering\arraybackslash}m{0.11\linewidth}
>{\centering\arraybackslash}m{0.12\linewidth}
>{\centering\arraybackslash}m{0.08\linewidth}
>{\centering\arraybackslash}m{0.08\linewidth}
>{\centering\arraybackslash}m{0.08\linewidth}
}
\toprule
\textbf{Dataset} & \textbf{Method} & \textbf{OA (\%)} & \textbf{AA (\%)} & \textbf{Kappa ($\times 100$)} & \textbf{Total} & \textbf{Train} & \textbf{Upload} \\
\midrule

\multirow{6}{*}{SA}
 & NFT      & 92.12 & 94.79 & 91.24 & 273.1k & 0      & 0      \\
 & FT       & 92.85 & 95.18 & 92.04 & 273.1k & 273.1k & 273.1k \\
 & LoRA     & 93.33 & 95.78 & 92.57 & 324.7k & 51.5k  & 51.5k  \\
 & LoRA+    & 93.75 & 96.18 & 93.04 & 324.7k & 51.5k  & 51.5k  \\
 & VERA     & 88.03 & 86.51 & 86.63 & 325.0k & 0.4k   & 0.4k   \\
 & \textbf{NE-LoRA} & \textbf{94.35 (+2.33)} & \textbf{96.53 (+2.84)} & \textbf{93.71 (+2.61)} & \textbf{376.2k} & \textbf{103.1k} & \textbf{103.1k (2.6$\times$)} \\
\midrule

\multirow{6}{*}{PU}
 & NFT      & 93.90 & 91.09 & 91.89 & 81.9k  & 0      & 0      \\
 & FT       & 95.49 & 95.49 & 94.05 & 81.9k  & 81.9k  & 81.9k  \\
 & LoRA     & 96.13 & 95.11 & 94.87 & 107.3k & 25.4k  & 25.4k  \\
 & LoRA+    & 95.92 & 95.34 & 94.61 & 107.3k & 25.4k  & 25.4k  \\
 & VERA     & 86.51 & 82.93 & 81.54 & 107.7k & 0.4k   & 0.4k   \\
 & \textbf{NE-LoRA} & \textbf{97.06 (+3.47)} & \textbf{96.42 (+4.43)} & \textbf{96.11 (+4.72)} & \textbf{132.7k} & \textbf{50.8k} & \textbf{25.4k (3.2$\times$)} \\
\midrule

\multirow{6}{*}{HR-L}
 & NFT      & 77.26 & 66.38 & 72.54 & 208.6k & 0      & 0      \\
 & FT       & 82.04 & 73.07 & 78.46 & 208.6k & 208.6k & 208.6k \\
 & LoRA     & 82.08 & 73.49 & 78.57 & 253.2k & 44.6k  & 44.6k  \\
 & LoRA+    & 82.63 & 74.79 & 79.39 & 253.2k & 44.6k  & 44.6k  \\
 & VERA     & 71.82 & 55.82 & 65.96 & 253.6k & 0.4k   & 0.4k   \\
 & \textbf{NE-LoRA} & \textbf{84.57 (+5.40)} & \textbf{77.39 (+8.68)} & \textbf{81.63 (+6.65)} & \textbf{297.8k} & \textbf{89.2k } & \textbf{44.6k (4.7$\times$)} \\
\midrule

\multirow{6}{*}{WHU-LK}
 & NFT      & 89.83 & 63.41 & 86.50 & 210.9k & 0      & 0      \\
 & FT       & 91.30 & 63.80 & 88.39 & 210.9k & 210.9k & 210.9k \\
 & LoRA     & 91.21 & 64.69 & 88.28 & 279.3k & 68.4k  & 68.4k  \\
 & LoRA+    & 91.19 & 64.90 & 88.26 & 279.3k & 68.4k  & 68.4k  \\
 & VERA     & 89.61 & 59.45 & 86.18 & 279.7k & 0.4k   & 0.4k   \\
 & \textbf{NE-LoRA} & \textbf{91.85 (+1.22)} & \textbf{66.87 (+3.62)} & \textbf{89.11 (+1.59)} & \textbf{347.7k} & \textbf{136.8k} & \textbf{68.4k (3.1$\times$)} \\
\bottomrule
\end{tabular}
\end{table}

\begin{table}[t]
\centering
\setlength{\tabcolsep}{2.5pt}
\renewcommand{\arraystretch}{0.9}
\scriptsize
\caption{Performance of different methods on HybridSN.
Numbers in parentheses indicate the average improvement of NE-LoRA over the compared methods for each column.}
\label{tab:hybridsn_performance}
\begin{tabular}{
>{\centering\arraybackslash}m{0.06\linewidth}
>{\centering\arraybackslash}m{0.11\linewidth}
>{\centering\arraybackslash}m{0.11\linewidth}
>{\centering\arraybackslash}m{0.11\linewidth}
>{\centering\arraybackslash}m{0.12\linewidth}
>{\centering\arraybackslash}m{0.08\linewidth}
>{\centering\arraybackslash}m{0.08\linewidth}
>{\centering\arraybackslash}m{0.08\linewidth}
}
\toprule
\textbf{Dataset} & \textbf{Method} & \textbf{OA (\%)} & \textbf{AA (\%)} & \textbf{Kappa ($\times 100$)} & \textbf{Total} & \textbf{Train} & \textbf{Upload} \\
\midrule

\multirow{6}{*}{SA}
 & NFT      & 85.53 & 67.33 & 83.83 & 452.7k & 0      & 0      \\
 & FT       & \textbf{92.79} & \textbf{82.92} & \textbf{91.96} & 452.7k & 452.7k & 452.7k \\
 & LoRA     & 89.98 & 74.02 & 88.82 & 463.9k & 11.2k  & 11.2k  \\
 & LoRA+    & 89.21 & 73.56 & 87.96 & 463.9k & 11.2k  & 11.2k  \\
 & VERA     & 86.78 & 74.45 & 85.26 & 476.5k & 1.3k   & 1.3k   \\
 & NE-LoRA  & 90.84 (+1.98) & 80.12 (+5.66) & 89.80 (+2.23) & \textbf{475.1k} & \textbf{22.4k} & \textbf{11.2k (40.4$\times$)} \\
\midrule

\multirow{6}{*}{PU}
 & NFT      & 95.64 & 92.04 & 94.21 & 451.8k & 0      & 0      \\
 & FT       & 97.34 & 94.68 & 96.47 & 451.8k & 451.8k & 451.8k \\
 & LoRA     & 94.31 & 85.37 & 92.43 & 463.0k & 11.2k  & 11.2k  \\
 & LoRA+    & 97.05 & 94.71 & 96.08 & 463.0k & 11.2k  & 11.2k  \\
 & VERA     & 87.09 & 75.55 & 82.78 & 464.3k & 1.3k   & 1.3k   \\
 & \textbf{NE-LoRA} & \textbf{97.43 (+3.14)} & \textbf{95.07 (+6.60)} & \textbf{96.59 (+4.20)} & \textbf{474.2k} & \textbf{22.3k} & \textbf{11.2k (40.3$\times$)} \\
\midrule

\multirow{6}{*}{HR-L}
 & NFT      & 78.96 & 62.38 & 74.86 & 452.5k & 0      & 0      \\
 & FT       & \textbf{86.83} & \textbf{74.33} & \textbf{84.30} & 452.5k & 452.5k & 452.5k \\
 & LoRA     & 84.07 & 70.10 & 81.05 & 463.7k & 11.2k  & 11.2k  \\
 & LoRA+    & 85.69 & 77.11 & 82.96 & 463.7k & 11.2k  & 11.2k  \\
 & VERA     & 72.03 & 50.88 & 66.34 & 465.0k & 1.3k   & 1.3k   \\
 & NE-LoRA  & 84.79 (+3.27) & 68.39 (+1.43) & 81.87 (+3.97) & \textbf{474.9k} & \textbf{22.4k} & \textbf{11.2k (40.4$\times$)} \\
\midrule

\multirow{6}{*}{WHU-LK}
 & NFT      & 80.97 & 54.99 & 74.45 & 451.8k & 0      & 0      \\
 & FT       & 88.83 & 68.10 & 85.23 & 451.8k & 451.8k & 451.8k \\
 & LoRA     & 83.45 & 61.38 & 78.08 & 463.0k & 11.2k  & 11.2k  \\
 & LoRA+    & 85.17 & 61.60 & 80.23 & 463.0k & 11.2k  & 11.2k  \\
 & VERA     & 84.96 & 51.66 & 79.63 & 464.3k & 1.3k   & 1.3k   \\
 & \textbf{NE-LoRA} & \textbf{89.74 (+5.06)} & \textbf{68.13 (+8.58)} & \textbf{86.45 (+6.93)} & \textbf{474.2k} & \textbf{22.3k} & \textbf{11.2k (40.4$\times$)} \\
\bottomrule
\end{tabular}
\end{table}
\subsection{Stability and Robustness Analysis}
\begin{figure*}[t]
    \centering
    
    % 上方图例
    \includegraphics[width=0.35\textwidth]{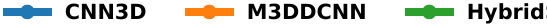}
    \vspace{0.15cm}
    
    % 四张图一行
    \begin{minipage}[b]{0.24\textwidth}
        \centering
        \includegraphics[width=\linewidth]{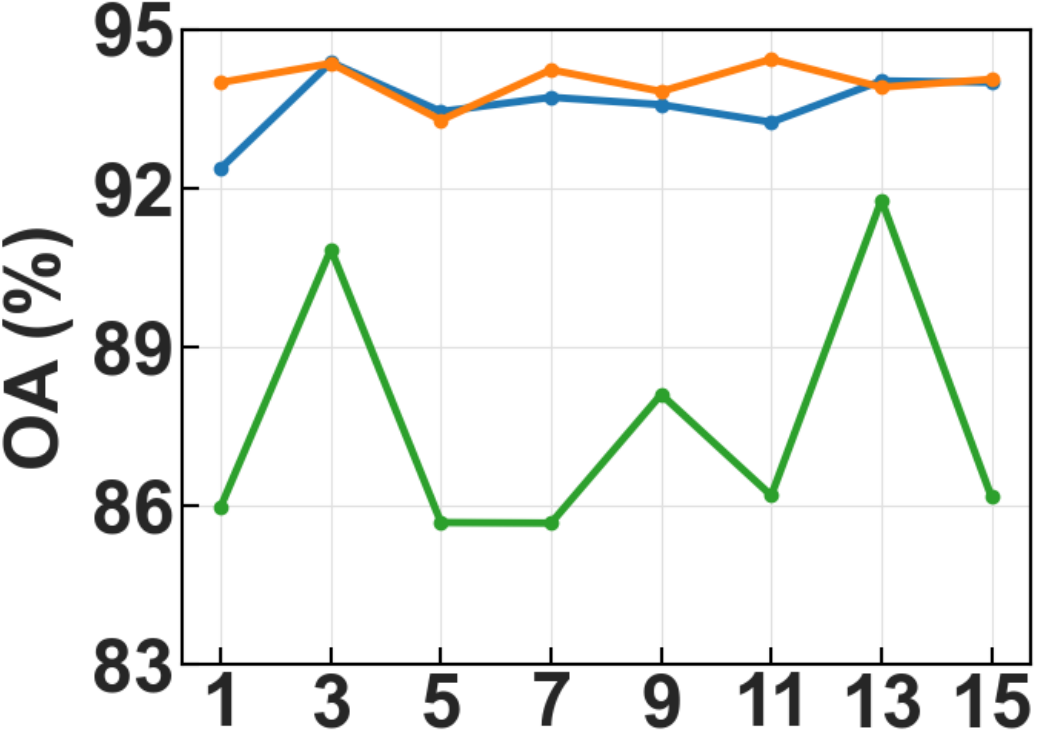}
        \vspace{-0.1cm}
        
        \small\textbf{SA}
    \end{minipage}
    \hfill
    \begin{minipage}[b]{0.24\textwidth}
        \centering
        \includegraphics[width=\linewidth]{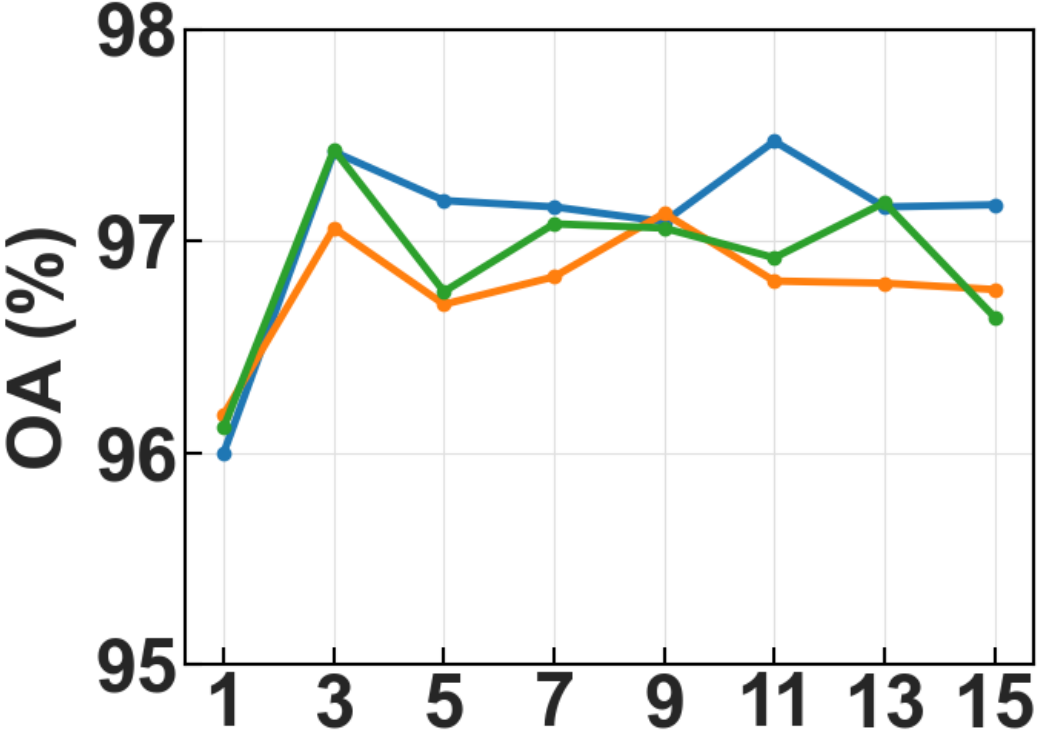}
        \vspace{-0.1cm}
        
        \small\textbf{PU}
    \end{minipage}
    \hfill
    \begin{minipage}[b]{0.24\textwidth}
        \centering
        \includegraphics[width=\linewidth]{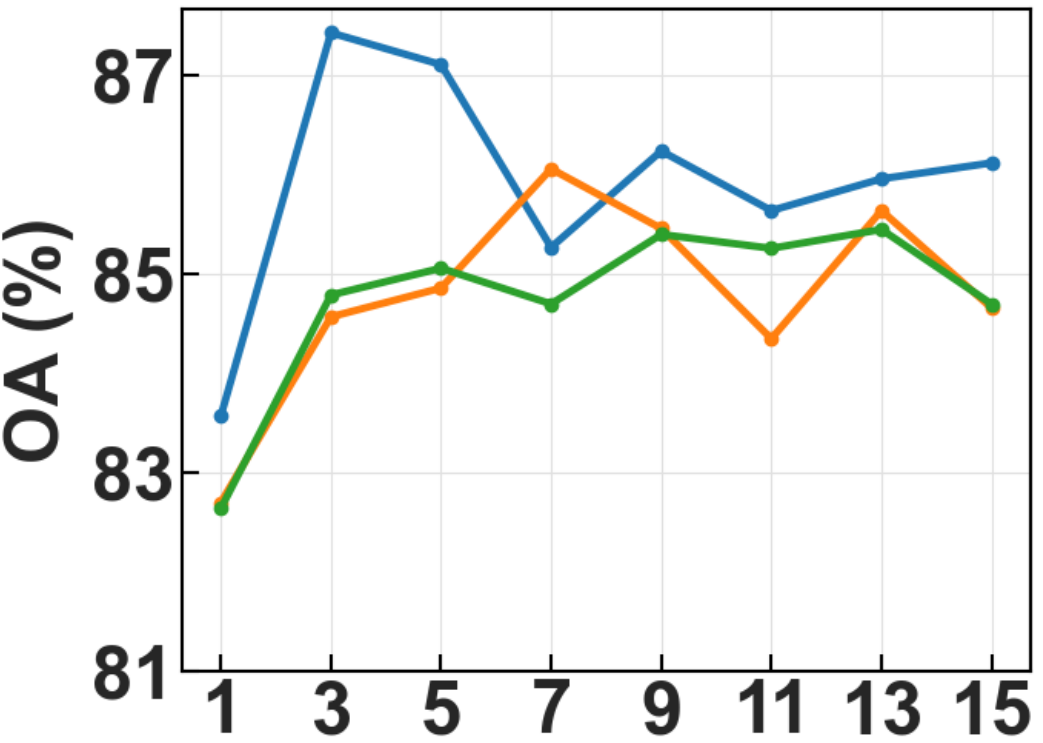}
        \vspace{-0.1cm}
        
        \small\textbf{HR-L}
    \end{minipage}
    \hfill
    \begin{minipage}[b]{0.24\textwidth}
        \centering
        \includegraphics[width=\linewidth]{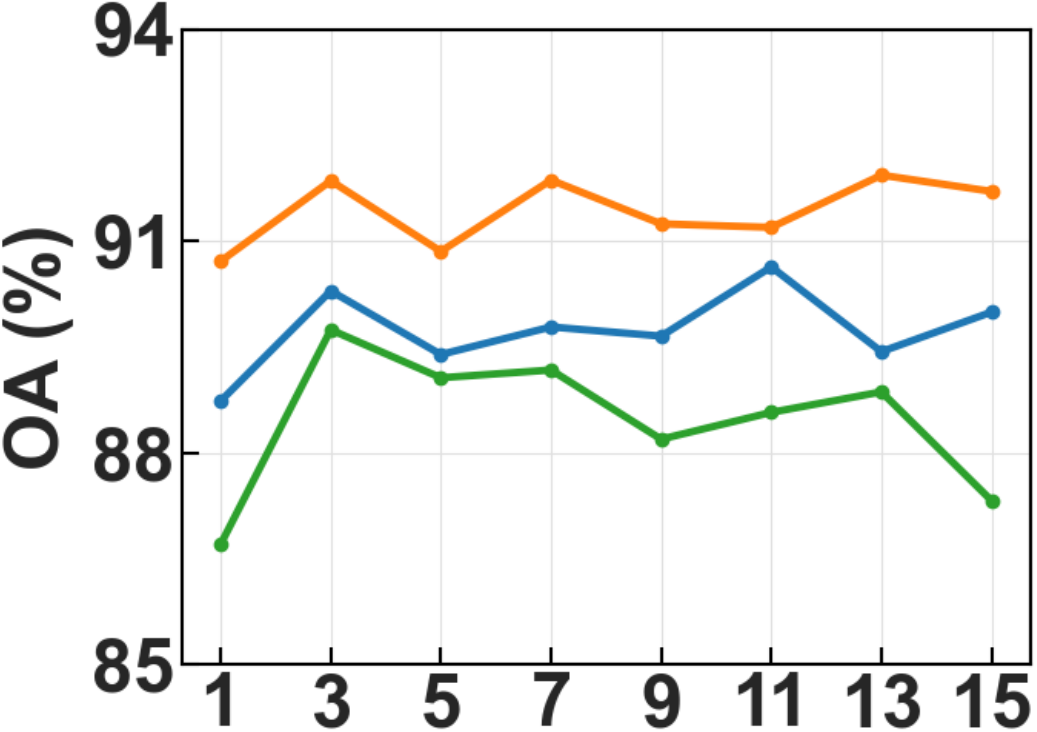}
        \vspace{-0.1cm}
        
        \small\textbf{WHU-LK}
    \end{minipage}
    
    \caption{Accuracy variation of models across datasets with varying rank.}
    \label{fig:model_accuracy_rank}
    \vspace{-0.5em}
\end{figure*}

\textbf{Robustness to Low-Rank Constraints.} We first investigate the stability of NE-LoRA under varying low-rank constraints $r$.
As illustrated in Fig. \ref{fig:model_accuracy_rank},  the classification accuracy saturates at a very small rank (typically $r=3$) and remains consistently high thereafter. This trend reinforces the intrinsic low-rank property of hyperspectral features, suggesting that essential spectral–spatial patterns can be encapsulated within a minimal parameter budget.
Notably, even in the extreme case of $r=1$, NE-LoRA maintains high-level semantic capture and competitive accuracy. This serves as a strong validation that our nonlinear auxiliary branch (GELU) effectively compensates for the limited representational capacity of purely linear adapters, allowing for "High Performance" even under the most stringent "Low Cost" constraints

\begin{figure}[t]
    \centering
    \includegraphics[width=\linewidth]{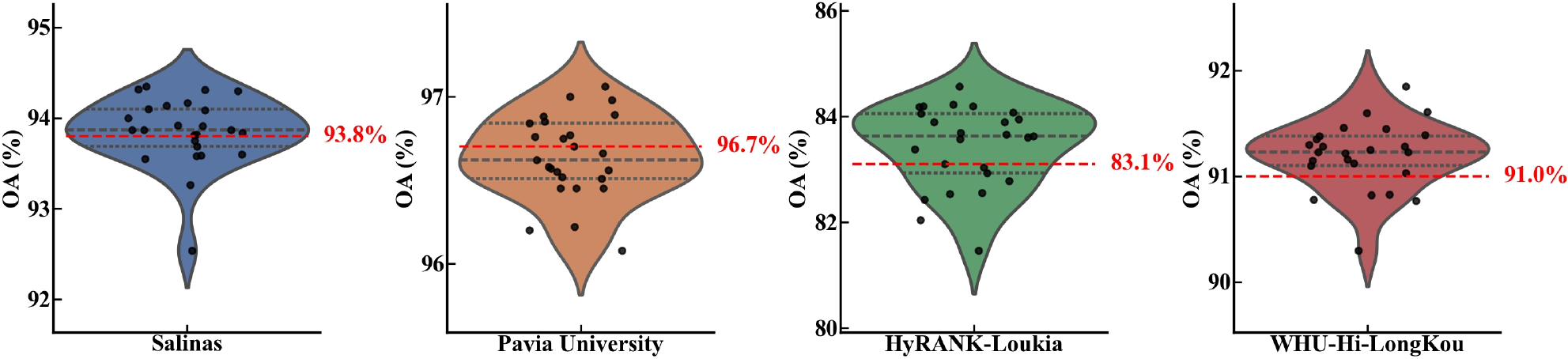}
    \caption{Accuracy distribution across datasets with fixed $\lambda$ and variable $\gamma$ for large matrices.}
    \label{fig:violin_large_matrix}
    \vspace{-0.5cm}
\end{figure}

\textbf{Robustness to Optimization Hyper-parameters.} Next, we erform a robustness validation of the learning rate modulation factor $\gamma$ for large-proportion modules.
Fig. \ref{fig:violin_large_matrix} presents the accuracy distribution across a wide range of $\gamma$ setting, with the red dashed line representing the baseline without modulation.
The results show that performance remains at or above the baseline across nearly all evaluated $\gamma$ values.
This parameter insensitivity is vital for LEO satellite nodes, where limited onboard energy and intermittent connectivity make frequent, model-specific fine-tuning infeasible. By moderating input-side updates, $\gamma$ prevents aggressive optimization and unstable gradient dynamics in high-dimensional modules while preserving sufficient adaptation capacity. Overall, these findings confirm that NE-LoRA is highly robust to hyper-parameter selection, ensuring reliable model evolution in dynamic orbital environments.

\newpage

\section*{NeurIPS Paper Checklist}

%%% BEGIN INSTRUCTIONS %%%
The checklist is designed to encourage best practices for responsible machine learning research, addressing issues of reproducibility, transparency, research ethics, and societal impact. Do not remove the checklist: {\bf The papers not including the checklist will be desk rejected.} The checklist should follow the references and follow the (optional) supplemental material.  The checklist does NOT count towards the page
limit. 

Please read the checklist guidelines carefully for information on how to answer these questions. For each question in the checklist:
\begin{itemize}
    \item You should answer \answerYes{}, \answerNo{}, or \answerNA{}.
    \item \answerNA{} means either that the question is Not Applicable for that particular paper or the relevant information is Not Available.
    \item Please provide a short (1--2 sentence) justification right after your answer (even for \answerNA). 
   % \item {\bf The papers not including the checklist will be desk rejected.}
\end{itemize}

{\bf The checklist answers are an integral part of your paper submission.} They are visible to the reviewers, area chairs, senior area chairs, and ethics reviewers. You will also be asked to include it (after eventual revisions) with the final version of your paper, and its final version will be published with the paper.

The reviewers of your paper will be asked to use the checklist as one of the factors in their evaluation. While \answerYes{} is generally preferable to \answerNo{}, it is perfectly acceptable to answer \answerNo{} provided a proper justification is given (e.g., error bars are not reported because it would be too computationally expensive'' or ``we were unable to find the license for the dataset we used''). In general, answering \answerNo{} or \answerNA{} is not grounds for rejection. While the questions are phrased in a binary way, we acknowledge that the true answer is often more nuanced, so please just use your best judgment and write a justification to elaborate. All supporting evidence can appear either in the main paper or the supplemental material, provided in appendix. If you answer \answerYes{} to a question, in the justification please point to the section(s) where related material for the question can be found.

IMPORTANT, please:
\begin{itemize}
    \item {\bf Delete this instruction block, but keep the section heading ``NeurIPS Paper Checklist"},
    \item  {\bf Keep the checklist subsection headings, questions/answers and guidelines below.}
    \item {\bf Do not modify the questions and only use the provided macros for your answers}.
\end{itemize}

%%% END INSTRUCTIONS %%%

\begin{enumerate}

\item {\bf Claims}
    \item[] Question: Do the main claims made in the abstract and introduction accurately reflect the paper's contributions and scope?
    \item[] Answer: \answerYes{} % Replace by \answerYes{}, \answerNo{}, or \answerNA{}.
    \item[] Justification: Our contribution is outlined as a seperated paragraph in §1.

\item {\bf Limitations}
    \item[] Question: Does the paper discuss the limitations of the work performed by the authors?
    \item[] Answer: \answerYes{} % Replace by \answerYes{}, \answerNo{}, or \answerNA{}.
    \item[] Justification: We thoroughly discuss the limitations of our work in §6.

\item {\bf Theory assumptions and proofs}
    \item[] Question: For each theoretical result, does the paper provide the full set of assumptions and a complete (and correct) proof?
    \item[] Answer: \answerNA{} % Replace by \answerYes{}, \answerNo{}, or \answerNA{}.
    \item[] Justification: This paper does not include theoretical results.

    \item {\bf Experimental result reproducibility}
    \item[] Question: Does the paper fully disclose all the information needed to reproduce the main experimental results of the paper to the extent that it affects the main claims and/or conclusions of the paper (regardless of whether the code and data are provided or not)?
    \item[] Answer: \answerYes{} % Replace by \answerYes{}, \answerNo{}, or \answerNA{}.
    \item[] Justification: We provide detailed instructions on how to reproduce the main experimental
results in §5. We will open-source the code and data upon acceptance.

\item {\bf Open access to data and code}
    \item[] Question: Does the paper provide open access to the data and code, with sufficient instructions to faithfully reproduce the main experimental results, as described in supplemental material?
    \item[] Answer: \answerNo{} % Replace by \answerYes{}, \answerNo{}, or \answerNA{}.
    \item[] Justification: We will open-source the code and data upon acceptance.

\item {\bf Experimental setting/details}
    \item[] Question: Does the paper specify all the training and test details (e.g., data splits, hyperparameters, how they were chosen, type of optimizer) necessary to understand the results?
    \item[] Answer: \answerYes{} % Replace by \answerYes{}, \answerNo{}, or \answerNA{}.
    \item[] Justification:  We provide detailed instructions on how to reproduce the main experimental results in §4.

\item {\bf Experiment statistical significance}
    \item[] Question: Does the paper report error bars suitably and correctly defined or other appropriate information about the statistical significance of the experiments?
    \item[] Answer: \answerNo{} % Replace by \answerYes{}, \answerNo{}, or \answerNA{}.
    \item[] Justification: Error bars are not reported because of the time limit. We will attempt to add
them in the camera-ready version.


\item {\bf Experiments compute resources}
    \item[] Question: For each experiment, does the paper provide sufficient information on the computer resources (type of compute workers, memory, time of execution) needed to reproduce the experiments?
    \item[] Answer: \answerYes{} % Replace by \answerYes{}, \answerNo{}, or \answerNA{}.
    \item[] Justification: We provide detailed hardware information in §4.
    
\item {\bf Code of ethics}
    \item[] Question: Does the research conducted in the paper conform, in every respect, with the NeurIPS Code of Ethics \url{https://neurips.cc/public/EthicsGuidelines}?
    \item[] Answer: \answerYes{} % Replace by \answerYes{}, \answerNo{}, or \answerNA{}.
    \item[] Justification: We have reviewed the NeurIPS Code of Ethics and believe that our research conforms to it.

\item {\bf Broader impacts}
    \item[] Question: Does the paper discuss both potential positive societal impacts and negative societal impacts of the work performed?
    \item[] Answer: \answerYes{} % Replace by \answerYes{}, \answerNo{}, or \answerNA{}.
    \item[] Justification: We have discussed and provided real-world examples of both positive and negative societal impacts in §1 and §2.
    
\item {\bf Safeguards}
    \item[] Question: Does the paper describe safeguards that have been put in place for responsible release of data or models that have a high risk for misuse (e.g., pre-trained language models, image generators, or scraped datasets)?
    \item[] Answer: \answerNA{} % Replace by \answerYes{}, \answerNo{}, or \answerNA{}.
    \item[] Justification: This paper is intended for onboard model adaptation and does not involve high-risk data or models.

\item {\bf Licenses for existing assets}
    \item[] Question: Are the creators or original owners of assets (e.g., code, data, models), used in the paper, properly credited and are the license and terms of use explicitly mentioned and properly respected?
    \item[] Answer: \answerYes{} % Replace by \answerYes{}, \answerNo{}, or \answerNA{}.
    \item[] Justification: We have properly cited the original data and models in §4.

\item {\bf New assets}
    \item[] Question: Are new assets introduced in the paper well documented and is the documentation provided alongside the assets?
    \item[] Answer: \answerNo{} % Replace by \answerYes{}, \answerNo{}, or \answerNA{}.
    \item[] Justification: We will provide detailed documentation for the new assets upon acceptance.

\item {\bf Crowdsourcing and research with human subjects}
    \item[] Question: For crowdsourcing experiments and research with human subjects, does the paper include the full text of instructions given to participants and screenshots, if applicable, as well as details about compensation (if any)? 
    \item[] Answer: \answerNA{} % Replace by \answerYes{}, \answerNo{}, or \answerNA{}.
    \item[] Justification:  This paper does not involve crowdsourcing nor research with human subjects.

\item {\bf Institutional review board (IRB) approvals or equivalent for research with human subjects}
    \item[] Question: Does the paper describe potential risks incurred by study participants, whether such risks were disclosed to the subjects, and whether Institutional Review Board (IRB) approvals (or an equivalent approval/review based on the requirements of your country or institution) were obtained?
    \item[] Answer: \answerNA{} % Replace by \answerYes{}, \answerNo{}, or \answerNA{}.
    \item[] Justification:  This paper does not involve crowdsourcing nor research with human subjects.

\item {\bf Declaration of LLM usage}
    \item[] Question: Does the paper describe the usage of LLMs if it is an important, original, or non-standard component of the core methods in this research? Note that if the LLM is used only for writing, editing, or formatting purposes and does \emph{not} impact the core methodology, scientific rigor, or originality of the research, declaration is not required.
    %this research? 
    \item[] Answer: \answerNA{} % Replace by \answerYes{}, \answerNo{}, or \answerNA{}.
    \item[] Justification: LLMs are not part of the core methodology of this work. Any LLM usage was limited to writing or editing assistance and did not affect the scientific content, experiments, or originality of the research.

\end{enumerate}

\end{document}